\documentclass[10pt,twocolumn,letterpaper]{article}

\usepackage{cvpr}              

\usepackage[dvipsnames]{xcolor}

\definecolor{cvprblue}{rgb}{0.21,0.49,0.74}
\usepackage[pagebackref,breaklinks,colorlinks,citecolor=cvprblue]{hyperref}
\usepackage{tabularx}
\usepackage[export]{adjustbox}
\usepackage{bm}
\usepackage{gensymb}
\usepackage[nolist]{acronym}
\usepackage{nicematrix}

\begin{acronym}
    \acro{ROI}[ROI]{Region of Interest}
    \acro{FOV}[FOV]{Field of View}
    \acro{MAE}[MAE]{Mean Absolute Error}
\end{acronym}

\def\paperID{24} 
\def\confName{CVPR}
\def\confYear{2024}

\title{Table tennis ball spin estimation with an event camera}

\author{Thomas Gossard\thanks{equal contribution\\This research was funded by Sony AI.}\qquad Julian Krismer$^{*}$\qquad Andreas Ziegler\qquad Jonas Tebbe\qquad Andreas Zell\\
Cognitive Systems Group, University of Tübingen, Germany\\
{\tt\small thomas.gossard@uni-tuebingen.de}
}

\begin{document}
\maketitle
\begin{abstract}
Spin plays a pivotal role in ball-based sports.
%
Estimating spin becomes a key skill due to its impact on the ball's trajectory and bouncing behavior.
Spin cannot be observed directly, making it inherently challenging to estimate.
In table tennis, the combination of high velocity and spin renders traditional low frame rate cameras inadequate for quickly and accurately observing the ball's logo to estimate the spin due to the motion blur.
Event cameras do not suffer as much from motion blur, thanks to their high temporal resolution.
Moreover, the sparse nature of the event stream solves communication bandwidth limitations many frame cameras face.
%
To the best of our knowledge, we present the first method for table tennis spin estimation using an event camera.
We use ordinal time surfaces to track the ball and then isolate the events generated by the logo on the ball.
Optical flow is then estimated from the extracted events to infer the ball's spin.
%
%
We achieved a spin magnitude mean error of $10.7 \pm 17.3$ rps and a spin axis mean error of $32.9 \pm 38.2\degree$ in real time for a flying ball.
%
\end{abstract}    
\section{Introduction}\label{sec:intro}

\begin{figure}[t]
  \centering
   \includegraphics[width=0.9\linewidth]{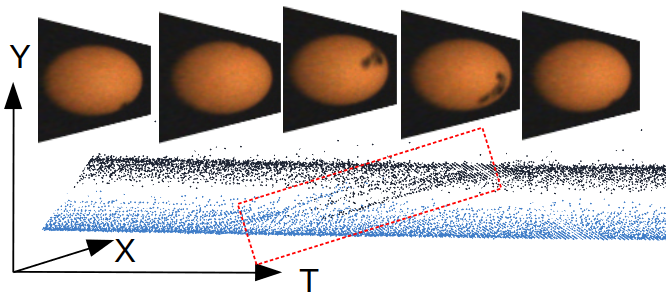}
   \caption{
        Comparison between the event stream of an event-based camera and captured frames from a frame-based camera.
        Black and blue dots are, respectively, ON- and OFF-events.
        The dotted red box highlights the events generated by the spinning logo.
   }
   \label{fig:events_vs_frames}
\end{figure}

Spin is crucial in ball sports, enhancing ball control and adding an element of unpredictability for the opponent. 
Whether it is topspin in tennis, backspin in table tennis, or spin variations in cricket, it adds complexity to the game. 
In soccer and baseball, players use spin to curve shots, making it challenging for opponents to predict the ball's trajectory.
This makes spin esimation decisive for playing such sports.
Thus, it is also a key component for developing autonomous agents and sports analysis systems.
However, in most cases, the spin is too fast for low frame rate cameras, which results in severe motion blur.
Event-based cameras do not suffer as much from motion blur as standard frame-based cameras.
Indeed, event cameras capture logarithmic changes in a scene's brightness with high temporal resolution in the order of $\mu$s.
Unlike traditional frame-based cameras, events are only emitted when there is a visual change, offering low latency and reduced bandwidth requirements.
This leads to much less information to be transmitted, given a scene with a static background.
As a consequence, we do not need to limit the \ac{ROI} of the camera to avoid bandwidth issues.
This was indeed a problem in~\cite{Tebbe2020icra}, where the full resolution of the camera ($1920$x$1200$ pixels) could not be used while maintaining a high enough frame rate ($380$ fps). 
Instead, a \ac{ROI} of $1920$x$400$ pixels was chosen.
The ball is thus observed for a shorter time, which decreases the likelihood of observing the ball's logo for low spin values.
The event camera can use its full resolution because it does not need to transmit redundant information.
The high temporal resolution of event cameras has an additional advantage. 
Estimating spin using a frame-based camera is constrained by the Nyquist-Shannon sampling theorem, which stipulates that the sampling rate must be at least twice the signal's bandwidth.
In other words, we need a camera with a certain frame rate to be able to measure high spin: $\text{fps} > |\omega| / 2$, where $\omega$ is the ball spin in $rps$.
%
In \cref{fig:events_vs_frames}, we can see that we only capture two frames where the logo is visible.
This makes it extremely difficult to estimate spin accurately.
On the other hand, the event camera could continuously capture the movement of the logo.
Occasionally, the ball rotates in a manner that causes the logo to consistently appear on the opposite side when captured by a frame-based camera.
This is no issue for event-based cameras, as events are generated asynchronously.
This overcomes the previously mentioned issue as long as the logo moves in front of the camera.
For all the previously mentioned reasons, event cameras are better suited for table tennis ball spin estimation.

\textbf{Contributions} of this work are as follows:

\begin{itemize}
    \item Tracking the ball in real-time and extracting events generated by the logo.
    \item A real-time and accurate method for table tennis ball spin estimation.
    \item An evaluation using a ball spinner to show the potential of the approach and a deployment with a ball thrower to show the approach in a real setup.
\end{itemize}
\section{Related work}\label{sec:related_work}

\subsection{Spin estimation}\label{subsec:rw_spin_estimation}

There are multiple approaches for spin estimation in table tennis.
Human players mostly estimate the spin applied to the ball from the motion of the opponent's racket.
This can be reproduced by using human pose estimation~\cite{Sato2020ceur,Kulkarni2021cvpr}, racket tracking~\cite{Gao2021icann}, or an inertial measurement unit (IMU) installed in the racket~\cite{Blank2017iswc,Blank2015iswc}.
However, the estimated spin is only an approximation, as there is no information on the racket's bounce dynamics, which are greatly influenced by its rubber type.
For a more quantitative spin estimate, there are two main approaches: observing the trajectory or directly observing the ball. 
Trajectory-based spin estimation~\cite{Wang2024tie,Su2013tim,Tebbe2020icra,Chen2010robio} makes use of the Magnus effect, which bends the ball's trajectory depending on the magnitude and type of spin applied.
Nevertheless, these approaches rely heavily on precise estimations of ball positions. 
The Magnus effect correlates directly with the level of spin imparted on the ball, complicating the differentiation between the spin-induced curvature of the trajectory and measurement inaccuracies of the ball positions.
While~\cite{Achterhold2023l4dc} showed that the ball's trajectory is insufficient for accurate spin estimation, the estimation of the spin can be further improved by using the bounce behavior of the ball~\cite{Wang2024tie}. 

However, for the most accurate spin estimation, ball observation methods are preferred.
Using the blur generated by the logo for spin estimation has been studied in~\cite{Boracchi2008cvta} but it does not seem to scale well to higher spin values.
Within the ball observation methods, we distinguish two sub-categories: logo-based~\cite{Glover2014icra,Tebbe2020icra,Zhang2015tim} and pattern-based~\cite{Theobalt2004siggraph,Szep2011cvww,Furuno2009iccas,Gossard2023iros,Tamaki2012icassp,Tamaki2004fcv}.
The patterns are always visible in contrast to the logo which is barely visible or hidden from a single camera's point of view, making previous logo-based approaches unreliable.
For official matches, custom balls are not allowed.
This makes logo-based methods necessary for competitive games.
Our proposed approach leverages event cameras for ball spin estimation.

All of the previously described methods are not exclusive.
They can be combined for better spin estimation.
A more recent work~\cite{Tamaki2024ijcss} combined a logo-based method with a trajectory-based approach.
The ambiguity about the spin magnitude stemming from the fact that the Nyquist-Shannon sampling theorem might not be fulfilled was resolved by estimating the ball trajectory for various spin magnitudes and choosing the spin magnitude for which the current ball trajectory fits best to the estimated one.

\subsection{Event cameras}\label{subsec:rw_event_cameras}

Event cameras offer many advantages, such as a high temporal resolution and a high dynamic range.
However, they also change the processing paradigm.
Most computer vision algorithms developed in the last decades are either ill-adapted to process events or do not make the most out of said advantages.
Frames can be reconstructed from events to apply state-of-the-art computer vision algorithms, but this introduces extra latency and does not fully utilize the event cameras' speed.

Tracking fast-moving objects is a perfect use case for event cameras, thanks to their high temporal resolution and less severe motion blur.
A lot of research has focused on fusing the data from frame-based and event-based cameras~\cite{Zhu2024arxiv,CamunasMesa2018tnnls,Chamorro2020bmvc,Chen2019icmm} to compensate for each other's weaknesses: frames for static or slow moving objects and events for fast-moving objects.

Because an event camera generates only ON/OFF events, new algorithms were developed to distinguish different moving objects in a scene.
\cite{Forrai2023icra,Falanga2020sr} tackle this issue by segmenting events generated by the background and events generated by independently moving objects with a motion-compensated mean time stamp image. 
Events are then clustered using DBSCAN.
Such methods were used for either obstacle avoidance for drones~\cite{Falanga2020sr} or for capturing a ball with a quadruped robot~\cite{Forrai2023icra}.
Next to the rather specific motion-compensated mean time stamp image, other common event representations are accumulated event frames and time-surfaces~\cite{Gallego2020pami}.
However, the same object will not be viewed the same way depending on its velocity.
This was solved with an ordinal time surface such as TOS~\cite{Glover2022pami} and EROS~\cite{Gava2022arxiv}.
These ordinal time surfaces capture the edge of objects without any trail of events, independent of the object's velocity. 
A similar representation could be achieved with accumulated event frames, but only if the accumulation time was tuned to fit the object's velocity.

Feature tracking can also benefit from using event cameras, as shown in~\cite{Alzugaray2020bmvc,Chiberre2021cvprw}.
However, because of the logo's small size and rapidly changing shape due to the curve of the spinning ball, such methods are difficult to apply.

Optical flow is an alternative for estimating the logo's speed. 
There are multiple ways to infer the optical flow.
There are Lucas-Kanade~\cite{Kanade1981iuw}, plane-fitting~\cite{Aung2018iscas}, and time surface matching~\cite{Nagata2021sensors}, just to name a few.
However, the current state-of-the-art is learning-based~\cite{Gehrig2024pami}. 
The most common architectures are U-Net~\cite{Ronneberger2015miccai}, FireNet~\cite{Scheerlinck2020wacv} and RAFT~\cite{Teed2020eccv}. 
The optical flow can be learned in either a supervised~\cite{Gehrig2024pami}, semi-supervised~\cite{Im2022eccv}, or unsupervised~\cite{Jonschkowski2020eccv} fashion.
Although more accurate, learning-based optical flow methods require more computational power, which makes achieving real-time performance difficult.
For this reason, we favor a model-based approach.

\section{Method}\label{sec:method}

Our table tennis ball spin estimation pipeline is divided into three phases.
Initially, a ball tracker, outlined in~\cref{subsec:ball_tracker}, estimates the ball's position, velocity, and radius.
Next, this information is used to extract events generated by the logo, as described in~\cref{subsec:extracting_logo}.
Finally, the ball's spin is estimated from the extracted events, detailed in~\cref{subsec:spin_estimation}.

\subsection{Ball tracker}\label{subsec:ball_tracker}

We set up a static event camera, observing a primarily static background (a table tennis table).
%
This configuration is beneficial since the flying ball triggers most of the events.

\begin{figure}[htb!]
  \centering
   \includegraphics[width=0.8\linewidth]{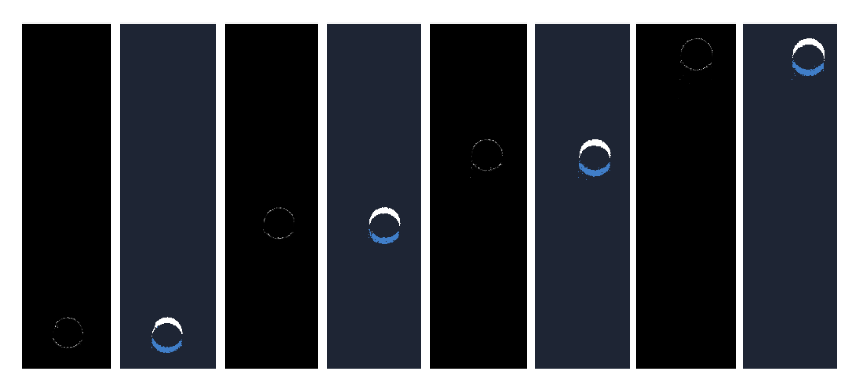}
   \caption{
        Pairs of representations for the same events. (left) EROS~\cite{Gava2022arxiv} event representations with $k_{eros}=10$.
        (right) Accumulated event frames with an accumulation time of $t_{acc}=2\text{ms}$.
   }
   \label{fig:eros_vs_acc}
\end{figure}

We use the Exponential Reduced Ordinal Surface (EROS)~\cite{Gava2022arxiv} event representation for processing the events.
EROS works by having new events decay the values of surrounding pixels by a factor, thus enabling objects to have sharp edges, whatever their velocity.
This representation enables continuous and asynchronous updates from the event stream, maximizing the potential of the event camera.
The update of EROS depends on a parameter named $k_{eros}$, which represents the update zone around the event.
$k_{eros}$ was hand-tuned for the best circle detection and set to $10$.
A comparison between the accumulated event frames and EROS can be seen in~\cref{fig:eros_vs_acc}.
One issue with the EROS time surface is the accumulation of noise.
Because the noise is spatially and temporally sparse, it stays in the representation. 
We thus “clean” the EROS time surface by removing isolated pixels.
We do so with a hit-or-miss transform to detect isolated non-zero pixels.
Detected isolated pixels are then set to zero. 
This hit-or-miss transform is performed on every update of the EROS time surface.
For more details, we refer to~\cref{sec:supp_cleaning_eros} in the supplementary material.

We do not use convolutions like in~\cite{Gava2022arxiv} because we also need the radius of the ball for extracting the events generated by the logo as explained in~\cref{subsec:spin_estimation}.
Instead, we apply a Hough circle transform to detect the ball and extract the peak value when the accumulation exceeds a certain threshold. 

\begin{figure}[htb!]
    \centering
    \includegraphics[width=\linewidth]{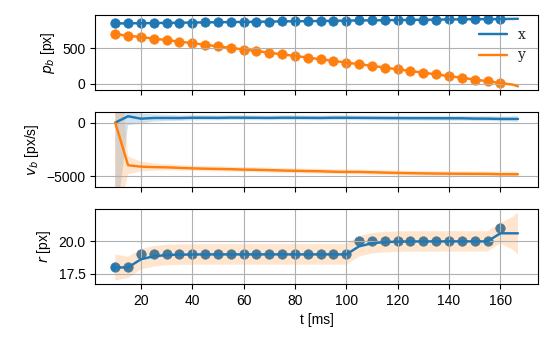}
    \caption{
            Tracked states of the Kalman filter: Positon, $\bm{p}_b = [x_b, y_b]$, velocity, $\bm{v}_b = [\dot{x}_b, \dot{y}_b]$ and ball radius $r$.
            The Kalman filter runs with $200$ Hz.
            The dots indicate measurements from the hough circle transform.
            The colored band around the estimated values represents the standard deviation.}
    \label{fig:kalman_filter}
\end{figure}

We added a Kalman filter to improve the ball tracker's performance.
This allows for smoother tracking as well as ball velocity estimation.
We also estimate the ball's radius to reduce the search space of the Hough circle transform for faster processing and for converting distances in pixels to meters. 
The state of the Kalman filter is defined with the state $\bm{\mathrm{x}}=[x_b, y_b, \dot{x}_b, \dot{y}_b, r]$ where $x_b$ and $y_b$ are the coordinates of the center of the ball, $\dot{x}_b$ and $\dot{y}_b$ are the ball's velocity in pixels/s and $r$ is the ball's radius in pixels.
Estimating the ball velocity and radius is relevant for extracting the events generated by the logo as explained in~\cref{subsec:extracting_logo}.
The state transition of the Kalman filter is defined as
\begin{equation}
    F = \begin{bmatrix}
    1 & 0 & dt & 0 & 0 \\
    0 & 1 & 0 & dt & 0 \\
    0 & 0 & 1 & 0 & 0 \\
    0 & 0 & 0 & 1 & 0 \\
    0 & 0 & 0 & 0 & 1 
    \end{bmatrix},
    \label{eq:F_mat}
\end{equation}
where $dt$ is the time difference between two Kalman filter updates.
The observation vector of the Kalman filter is $\bm{z} = [x_{b, \text{hough}}, y_{b, \text{hough}}, r_{\text{hough}}]$ where $x_{b, \text{hough}}$ and $y_{b, \text{hough}}$ are the coordinates of the ball center returned by the Hough circle transform and $r_{\text{hough}}$ is the corresponding radius.
The observation model of the Kalman filter is defined as
\begin{equation}
    H = \begin{bmatrix}
    1 & 0 & 0 & 0 & 0 \\
    0 & 1 & 0 & 0 & 0 \\
    0 & 0 & 0 & 0 & 1 
    \end{bmatrix}.
    \label{eq:H_mat}
\end{equation}
Though the ball's velocity is not directly observed, it can implicitly be estimated from the sequential measurements of ball positions.
The initial variance of the velocities must be set high enough to converge fast enough to the correct velocity.
In~\cref{fig:kalman_filter}, we show an example of the Kalman filter tracking the ball.
We can observe that the radius of the ball is not constant.
This is because the ball can get closer to the camera after being shot. 

\subsection{Extracting logo events}\label{subsec:extracting_logo}

The ball is tracked accurately using the previously described method.
With the Kalman filter, we have an estimate of the ball's velocity and radius.
The ball's velocity and radius can be used to extract the events generated “inside” the ball, i.e., events generated by the logo.
Events that we assume are generated by the logo have to satisfy
\begin{equation}
    (y_b(t_e) - y_e)^2 + (x_b(t_e) - x_e)^2 < (r - pad)^2,
    \label{eq:logo_cond}
\end{equation}
where $t_e$ is the event's time stamp, $x_e$ the x-positon of the event, $y_e$ the y-position of the event, $r$ the radius of the ball and $pad$ is a tolerance to avoid extracting events from the edge of the ball.

For maximal accuracy, the ball's position is continuously updated to the time stamp of the selected event, according to
\begin{equation}
    \begin{split}
     \bm{p_{b}}(t_e) &= (x_b(t_e), y_b(t_e)) \\
                      &= \bm{v_{b}}(t_i) \cdot (t_e - t_i) + \bm{p_{b}} (t_i),
    \end{split}
\label{eq:ball_pos}
\end{equation}
where $t_i$ is the previous time the Kalman filter's state was updated. 
With this, we can get the events generated by the logo of a flying ball.

\subsection{Spin Estimation}\label{subsec:spin_estimation}

Frame-based table tennis spin estimation methods capture successive logo orientations and then regress the spin. 
Though this method could be translated to event-based cameras, it would not fully exploit their advantages.

Generally, optical flow is a good tool for estimating velocities or, in this case, spin.
As mentioned in~\cref{sec:related_work}, there are numerous optical flow methods for event cameras.

\begin{figure}[htb!]
    \centering
    \includegraphics[width=\linewidth]{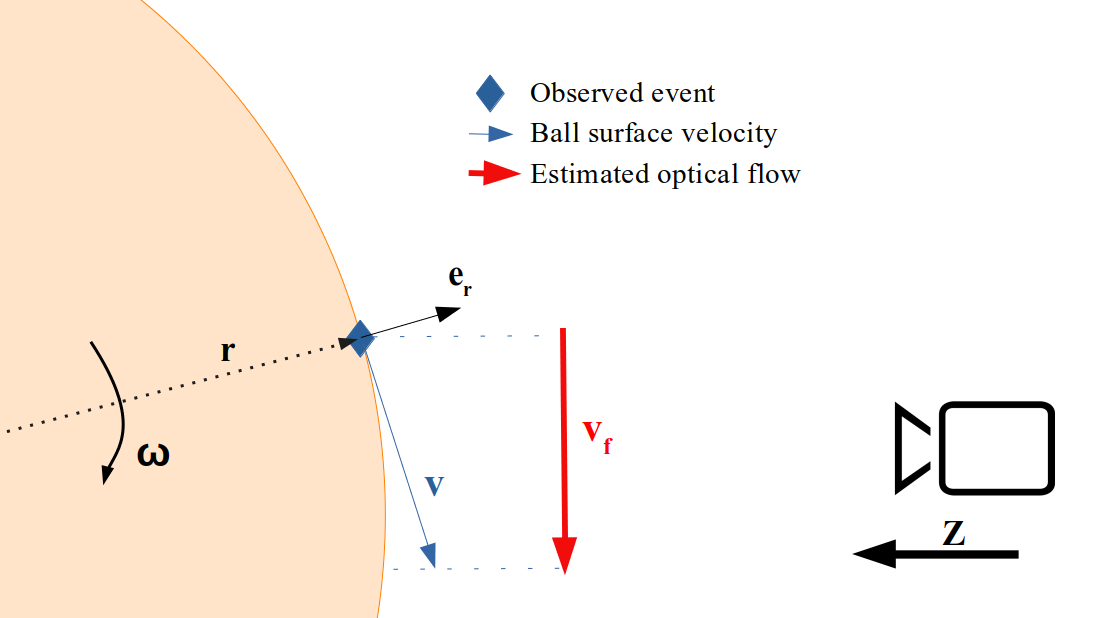}
    \caption{Sketch showing how the spin is calculated from the optical flow.}
    \label{fig:spin_inf_schematic}
\end{figure}

We first explain how we obtain the ball's spin $\bm{\omega}$ from the optical flow $v_f$ generated by the logo. 
From the flow, we can calculate the velocity $\bm{v}$ of the point on the ball's surface where the flow is generated
\begin{equation}
    \bm{v} = \begin{bmatrix}
           v_{f, x}\\
           v_{f, y} \\
           \frac{-v_{f,x} e_{r,x} - v_{f,y} e_{r,y}}{e_{r,z}}
         \end{bmatrix}.
\label{eq:v_flow}
\end{equation}
We can calculate the $\bm{z}$ component of $\bm{v}$ by knowing that the velocity is tangent to the ball's surface, $\bm{e_r}\cdot \bm{v} = 0$.
This is sketched in~\cref{fig:spin_inf_schematic}.
We can then calculate the rotation vector from the cross-product of the radius and the velocity.
\begin{equation}
    \bm{\omega} = \frac{\bm{e_r}\times\bm{v}}{||\bm{r}||}.
\end{equation}

When working with different optical flow methods, we noticed that they are able to infer flow correctly when the logo is in the center of the ball.
But they fail as soon as the logo is on the edge of the ball.
This is due to multiple factors.
First, table tennis ball logos have multiple edges.
As such, events are simultaneously generated at different places of the logo and not in a straight line.
This makes identifying successive events generated by the same edge difficult.
Secondly, because of the geometry of the ball, the logo's velocity is not constant.
There are several parameters to tune the different optical flow algorithms.
The main one is the accumulation time, $t_{acc}$, which aggregates events into an event frame used for optical flow.
To obtain the best optical flow results, the $t_{acc}$ should be chosen based on the logo's speed to ensure enough change to observe flow but not so much as to cause motion blur.
For a fast, first estimate, we choose $t_{acc}$ as a constant value, based on the median of the expected spin speeds as $t_{acc} = \lceil(\omega/10)\rceil$.
From testing, we divided each rotation into $10$ time slices because it ensures that even if the true spin speed is higher than we estimated, we still observe the flow multiple times per rotation.

\begin{figure}[htb!]
  \centering
   \includegraphics[width=\linewidth]{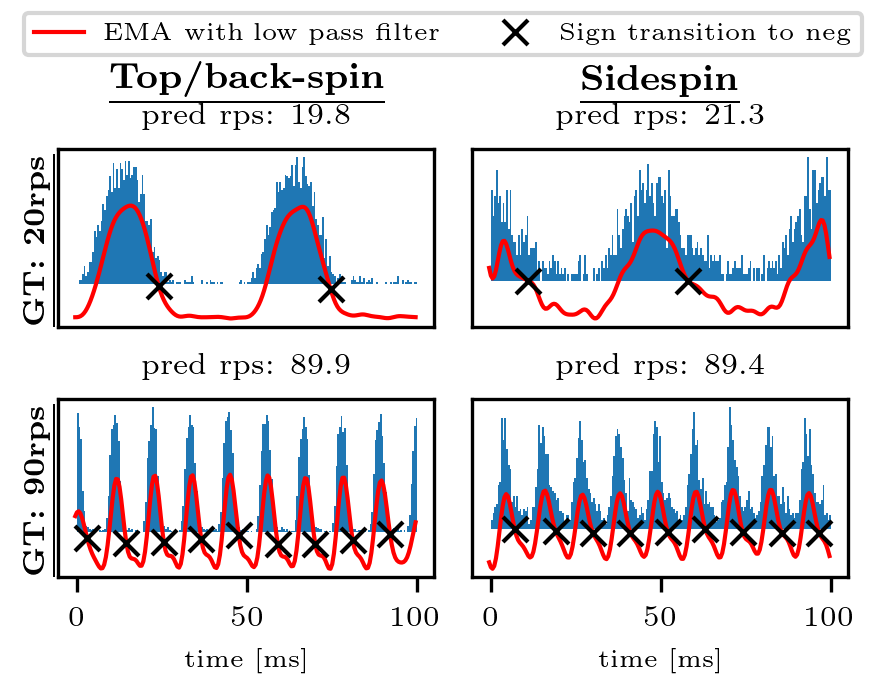}
   \caption{
        Prediction of the spin magnitude from the event rate. 
        The histograms of the event rate are displayed.
        We first apply an exponential moving average (EMA) and subtract the mean event rate.
        Then, we apply a low pass filter to smooth the curve.
        After that, we locate all the time stamps at which the sign of this curve transitions from positive to negative values.
    }
   \label{fig:ev_rate_pred}
\end{figure}

For a better estimate of the spin magnitude and therefore $t_{acc}$, we also consider a second approach based on the variation of the event rate from the extracted logo events.
Indeed, peaks in the event rate indicates when the logo is periodically the most visible.
This approach is inspired by~\cite{Pfrommer2022arxiv} and it returns results after a longer observation time but with a higher accuracy.
We first apply an exponential moving average (EMA) and subtract the mean event rate.
Then, we apply a low pass filter to smooth the curve.
After that, we locate all the time stamps at which the sign of this curve transitions from positive to negative values.
Since the change in the event rate is periodic and depends on the ball's spin, as the logo appears and disappears, it can be used to estimate the magnitude of the ball's spin.
\Cref{fig:ev_rate_pred} shows how the sign transitions of the event rates EMA can be used to calculate the magnitude of the spin, by taking the time difference between adjacent time stamps, calculating the mean, and converting this time per rotation into the number of rotations per second.

We can then apply the various optical flow methods for both methods of choosing $t_{acc}$.
The parameters are chosen for each one depending on $t_{acc}$ to return the best results.

For each $t_{acc}$, several flows and consequently spins are estimated, with the resulting value being the mean of all computed $\bm{\omega}$.
This process can be repeated for each $t_{acc}$ to generate additional spin estimations, which are then averaged collectively.

%
%
%




\section{Experiments}\label{sec:experiments}

We start this section by describing our experimental setup in~\cref{subsec:res_setup}.
In~\cref{subsec:res_ball_detection} we evaluate the ball detector and in~\cref{subsec:res_spin_estimation} we describe our experiment to evaluate the spin estimation of our proposed approach.

\subsection{Setup}\label{subsec:res_setup}

\begin{figure}[htb]
    \centering
    \includegraphics[width=0.8\linewidth]{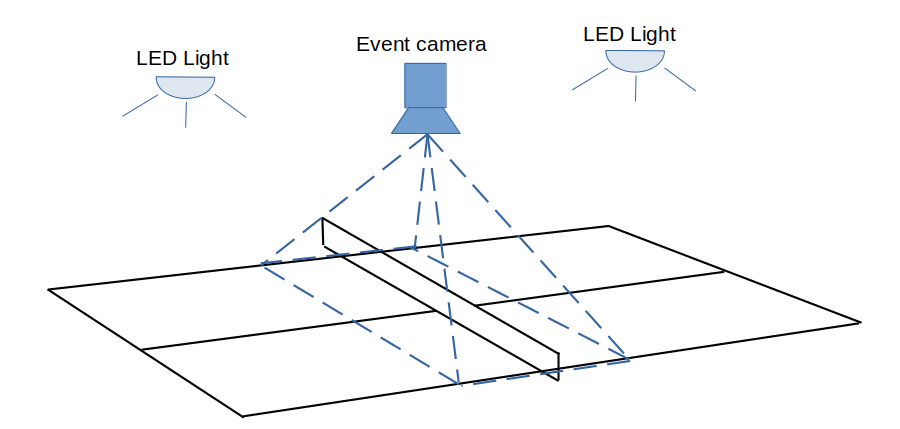}
    \caption{Event camera recording setup}
    \label{fig:camera_setup}
\end{figure}

We use a Prophesee EVK4 event camera with a resolution of $1280$x$720$ pixels for capturing events.
The camera is mounted $\sim 2\text{m}$ above the table tennis table, equipped with a $8\text{mm}$ lens.
This allows us to cover the whole width of the table, as shown in~\cref{fig:camera_setup}.
The camera's focus was adjusted using the protocol provided by Prophesee~\cite{focus_metavision}.
Event-based cameras possess particular settings that differ from standard, frame-based cameras.
These were investigated and tuned to generate the most events related to the ball and logo observation while minimizing noise.
For more information about the bias settings used with quantitative and qualitative results, please refer to~\cref{sec:res_camera_settings} in the supplementary material.
We use LED panels to light the scene uniformly from the ceiling, leading to a luminosity of around $1930\; \text{Lux}$ measured on the table.
This is to avoid the flickering light from standard lighting powered with $50\text{Hz}$ AC, which triggers events over the entire scene in the event camera, though events generated by the flickering light can also be filtered out with a band-cut filter.

\begin{figure}[htb!]
    \centering
    \includegraphics[width=0.9\linewidth]{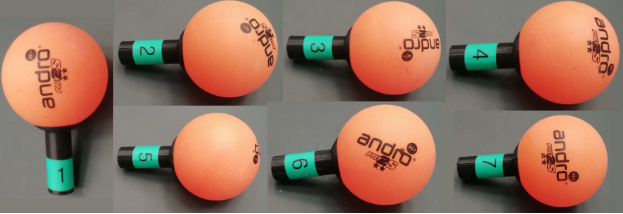}
    \caption{Balls with different logo orientations}
    \label{fig:balls}
\end{figure}

We rely on a custom-made static ball spinner and a ball thrower (Amicus Prime from Butterfly) to generate recordings with known ground truth.
The ground truth spin can be calculated from the ball spinner by knowing its relative position to the event camera.
We mounted the balls with different logo orientations on the ball spinner to study its impact on the spin estimation, as shown in~\cref{fig:balls}.

For the ball thrower, we rely on SpinDOE~\cite{Gossard2023iros} to generate ground truth spin, using balls with a custom pattern drawn on them.
A frame camera, the Grasshopper3 GS3-U3-23S6C (1900x400), was installed next to the event camera for capturing frames of the balls.
%





\subsection{Ball detection}\label{subsec:res_ball_detection}

\begin{figure}[htb!]
    \centering
    \includegraphics[width=\linewidth]{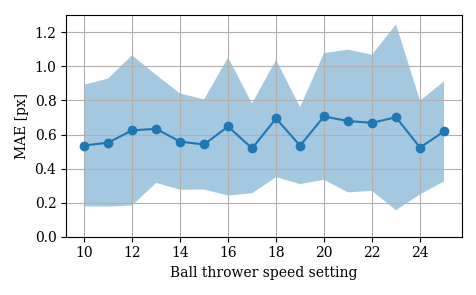}
    \caption{
            Evaluation of our ball detector for different ball thrower settings (integer values), reporting the \ac{MAE} in pixels.
            The colored band around the error represents the standard deviation.
    }
    \label{fig:ball_detect_bench}
\end{figure}

To evaluate the ball tracker, we recorded multiple balls flying in front of the event camera with speeds ranging from $4$ m/s to $12$ m/s. 
The ground truth was generated automatically with blob detection applied on accumulated event frames with a large accumulation time. 
In \cref{fig:ball_detect_bench}, we show our ball detector's \ac{MAE} for different ball thrower settings (arbitrary unit).
We can see that we have sub-pixel accuracy, which is necessary for extracting the events generated by the ball's logo. 

\begin{table*}[htb!]
    \centering
    \begin{NiceTabular}{*{2}{c}*{7}{c}{p{1.2cm}}}[hvlines]
    \CodeBefore
        \rectanglecolor{green!10}{3-3}{3-9}
        \rectanglecolor{green!10}{5-3}{5-9}
        \rectanglecolor{green!10}{7-3}{7-9}
        \rectanglecolor{green!10}{9-3}{9-9}
    \Body
                           & Ball                  & 1               & 2               & 3               & 4               & 6               & 7               \\
        \Block{4-1}{10ms}  & \Block{2-1}{Back/Top-spin}   & 30.0 $\pm$ 28.7 & 27.0 $\pm$ 27.5 & 23.36 $\pm$ 26.4 & 22.4 $\pm$ 21.9 & 32.0 $\pm$ 27.3 & 32.5 $\pm$ 28.2 \\
                           &                       & 18.1 $\pm$ 10.4 \degree & 25.3 $\pm$ 2.0 \degree & 27.6 $\pm$ 12.7 \degree & 25.6 $\pm$ 3.9 \degree & 15.4 $\pm$ 19.6 \degree & 9.5 $\pm$ 7.2 \degree \\
                           & \Block{2-1}{Sidespin} & -  & 31.0 $\pm$ 20.6 & 30.5 $\pm$ 18.8 & 32.0 $\pm$ 23.4 & - & -               \\
                           &                       & -               & 38.17 $\pm$  14.6 \degree & 40.3 $\pm$ 14.3 \degree & 42.8 $\pm$ 7.9 \degree & -               & -               \\
        \Block{4-1}{100ms} & \Block{2-1}{Back/Top-spin}  & 0.8 $\pm$ 0.1 & 0.5 $\pm$ 0.1 & 2.8 $\pm$ 2.7 & 3.7 $\pm$ 3.9 & 0.6 $\pm$ 0.1 & 0.5 $\pm$ 0.1 \\
                           &                       & 12.5 $\pm$ 4.0 \degree & 25.1 $\pm$ 1.8 \degree & 25.7 $\pm$ 1.4 \degree & 25.5 $\pm$ 1.9 \degree & 10.7 $\pm$ 5.5 \degree & 6.2 $\pm$ 3.2 \degree \\
                           & \Block{2-1}{Sidespin} & 4.8 $\pm$ 1.8 & 8.4 $\pm$ 1.5 & 8.0 $\pm$ 3.5 & 6.5 $\pm$ 3.2 & - & -               \\
                           &                       & -               & 16.7 $\pm$  9.4 \degree & 11.6 $\pm$ 13.5 \degree & 24.0 $\pm$ 10.5 \degree & -               & -               \\
    \end{NiceTabular}
    \caption{
            Evaluation of our spin estimation with the ball spinner.
            Our test was performed with balls with different logo positions relative to the spin axis, as shown in \cref{fig:balls} (white rows: MAE for the spin magnitude in rps, green rows: MAE for the spin axis in degrees).
            $10$ms and $100$ms indicate the amount of observation time allowed.
            Entries with "-" indicate that we did not get any results.
            For ball 1, the sidespin could not be estimated since the logo only rotates around the spin axis of the ball spinner, and there is no sidespin.
            The logo of balls 6 and 7 is a rotated version of ball 1; therefore, the same restriction applies.
    }
 \label{tab:results_by_ball}
\end{table*}
\begin{figure*}[htb!]
    \centering
    \includegraphics[width=\linewidth]{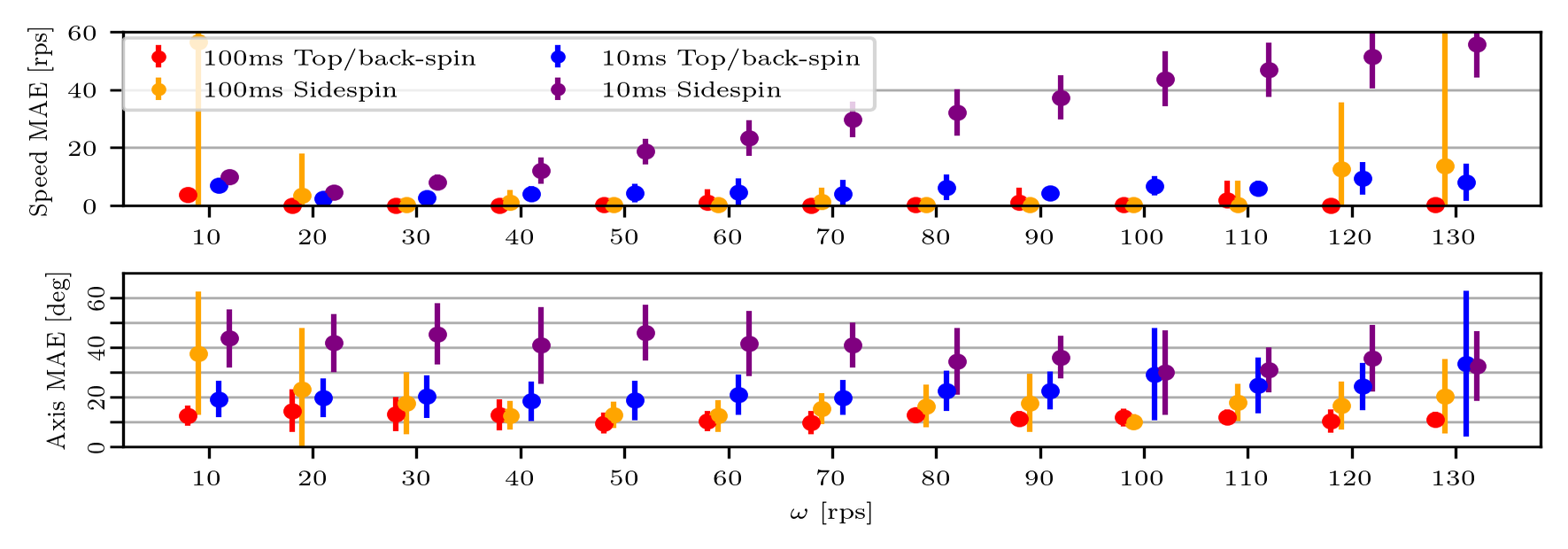}
    \caption{
            (top) Spin magnitude (speed) error with the ball spinner for different ball spins.
            (bottom) Spin axis estimation error with the ball spinner for different ball spins.
            $10$ms and $100$ms indicate the amount of observation time allowed.
            Only the events recorded during that time window can be used for the estimation.
            Backspin/Topspin was evaluated with balls 1, 6, and 7, and sidespin was evaluated with 2, 3, and 4, with an equal number of samples. 
    }
    \label{fig:results_by_rps}
\end{figure*}

\subsection{Spin estimation}\label{subsec:res_spin_estimation}

We tested our spin estimation method by two different means: with a ball spinner and with a ball thrower.
The former has the advantage to give us an extremely accurate and reliable ground truth while the later is closer to real use case.
Both benchmarks are still quite similar from the way events generated by the logo are extracted from the flying ball as explained in~\cref{subsec:extracting_logo}.
Indeed, the extracted events' position is recalculated using the ball position as a reference.

We selected Triplet Matching~\cite{Shiba2022lsp} as the optical flow algorithm.
Triplet Matching is the fastest optical flow method we found.
It has a sub-millisecond run-time while maintaining competitive accuracy.
Moreover, it runs on a CPU, which limits hardware requirements.

We distinguish two types of spin: backspin/topspin and sidespin. 
Because the ball is observed from the top, backspin, and topspin are almost the same, if only mirrored.
With top and backspin, the logo will come into view of the camera and then disappear again.
Sidespin balls, on the other hand, will either always have their logo hidden or always visible.
%

\subsubsection{Ball spinner benchmark}
We used table tennis balls with different logo orientations to evaluate our proposed spin estimation approach, as shown in~\cref{fig:balls}.
%
%
Next to the spin magnitude, our method also estimates the spin axis.
The spin magnitude estimation error and the spin axis estimation error with the ball spinner for different ball types are shown in~\cref{tab:results_by_ball}.
%

%
Entries with "-" indicate that we did not get any results.
For ball 1, the sidespin could not be estimated since the logo and spin axis coincide.
The logo of balls 6 and 7 is a rotated version of ball 1; therefore, the same restriction applies.

The errors with the ball spinner for different ball spins are reported in the top part of~\cref{fig:results_by_rps}.
As can be seen in the plot, the \ac{MAE} for the lowest spin magnitude of $10\text{rps}$ is very high.
The reason is twofold.
First, due to the relatively slow rotation speed, the logo triggers fewer events, which makes it difficult for the optical flow to get accurate flow estimations.
Second, with the relatively slow rotation speed, the likelihood of observing the logo over $10$ms is relatively low.
With $20$rps, the \ac{MAE} is already lower and relatively stable up to $100$rps for all four cases except for the sidespin with $10$ms observation time.
For sidespin with $10$ms observation time, the \ac{MAE} increases linearly with the spin speed.
This underlines the difficulty of estimating sidespin.
%

The \ac{MAE} of the spin axis estimation errors with the ball spinner for different ball spins are plotted in the bottom graph of~\cref{fig:results_by_rps}.
As can be seen in the plot, the spin axis estimation error decreases with higher spin values.
This is because higher spin values lead to more events, allowing the optical flow algorithm to achieve higher accuracy.





\subsubsection{Ball thrower benchmark}

Next to the evaluation with a ball spinner, we also deployed our method in a setup to estimate the spin of flying balls.
A ball thrower shot $20$ balls.
We used velocity settings of $10$, $15$, $20$, and $25$ for the ball thrower (respectively approximately $4$, $5.5$, $7.5$ and $9$ m/s ) and top- and side-spin, with spin strength settings ranging from $-5$ to $7$.
Both our event-based approach and a state-of-the-art frame-based approach~\cite{Tebbe2020icra} were tested on the same observations.
The results are listed in~\cref{tab:flying_ball_benchmark}.
\begin{table}[htb!]
    \centering
    \begin{tabular}{|l|l|l|}
        \hline
         & Frame & Event\\
         \hline
         Success rate& 0.79 & 0.81 \\
         \hline
         Spin axis MAE & $26.6 \pm 33.4\degree$ & $32.9 \pm 38.2\degree$ \\
         \hline
         Spin magnitude MAE & $8.8 \pm 15.6$ rps & $ 10.7\pm17.3$ rps \\
         \hline
    \end{tabular}
    \caption{Benchmark with the flying balls from a ball thrower}
    \label{tab:flying_ball_benchmark}
\end{table}
The success rate in the first row is the ratio between the number of successful spin estimations and the total number of ball trajectories.
As can be seen, our approach has a success rate slightly above the frame-based approach.
Our approach is a bit behind in terms of spin axis and spin magnitude \ac{MAE}, however, with overlapping error ranges.
Given that there is already a considerable body of literature covering frame-based spin estimation methods and our method is the first event-based one, this should show the potential of our approach.

\begin{figure*}[htb!]
    \centering
    \includegraphics[width=\linewidth]{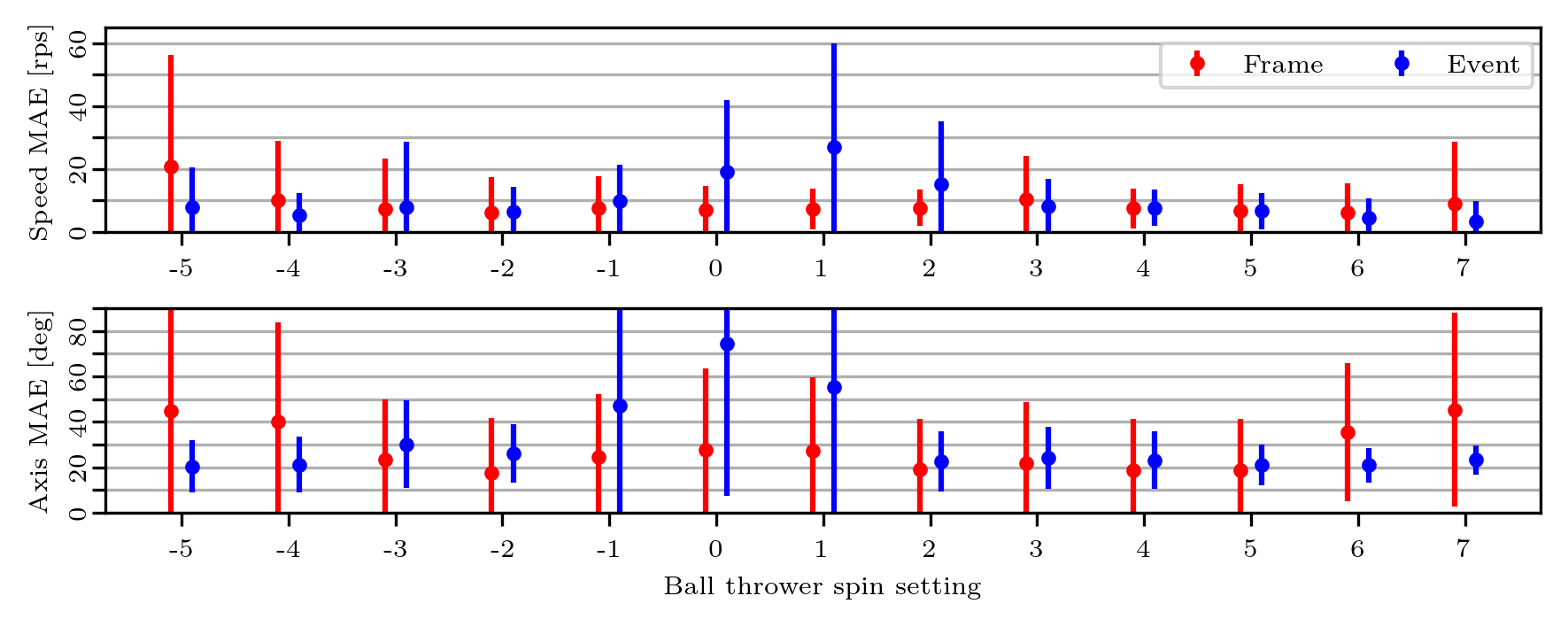}
    \caption{
            (top) Spin magnitude error with the ball thrower for different ball spins.
            (bottom) Spin axis estimation error with the ball thrower.
    }
    \label{fig:ball_thrower_benchmark}
\end{figure*}

In~\cref{fig:ball_thrower_benchmark}, we show the spin magnitude \ac{MAE} on the top and the spin axis \ac{MAE} on the bottom for a state-of-the-art frame-based approach~\cite{Tebbe2020icra} and our event-based approach.
As expected, we see that the frame-based approach has more problems with higher spin settings, while the event-based approach struggles with low spin settings.
It is worth noting that though both frame and event cameras have the same FOV, the frame camera boasts a higher resolution ($\sim \times 1.5$).
With a resolution similar to the frame camera, the event-based spin estimation would work better.
With these complementarity properties, an approach combining an event-based camera and a frame-based camera could be an option.

\subsection{Run times}

Since our goal is a real time capable spin estimation pipeline, the run-time is another important metric.
We used an Intel Core i7-9700 CPU @ 3.00GHz for our experiments, with all the components written in Python.

From the moment the events are received from the camera, generating the EROS time surface, detecting and tracking the ball, and extracting the logo events takes $12.8\pm1.2\text{ms}$.
We run the spin estimation in post-processing and report the ratio between the time it took to process all the events and the duration of the recording.
With a ratio of $0.84\pm0.8$, the spin estimation can process more data than it receives and, therefore, runs in real time.

While it can already be considered capable of being in real time, the setup could be improved.
Improving the software design and implementing the software pipeline in C++ instead of Python is expected to boost performance.

\subsection{Limitations}

\begin{figure}[htb!]
  \centering
    \begin{subfigure}{0.2\linewidth}
    \includegraphics[width=\linewidth]{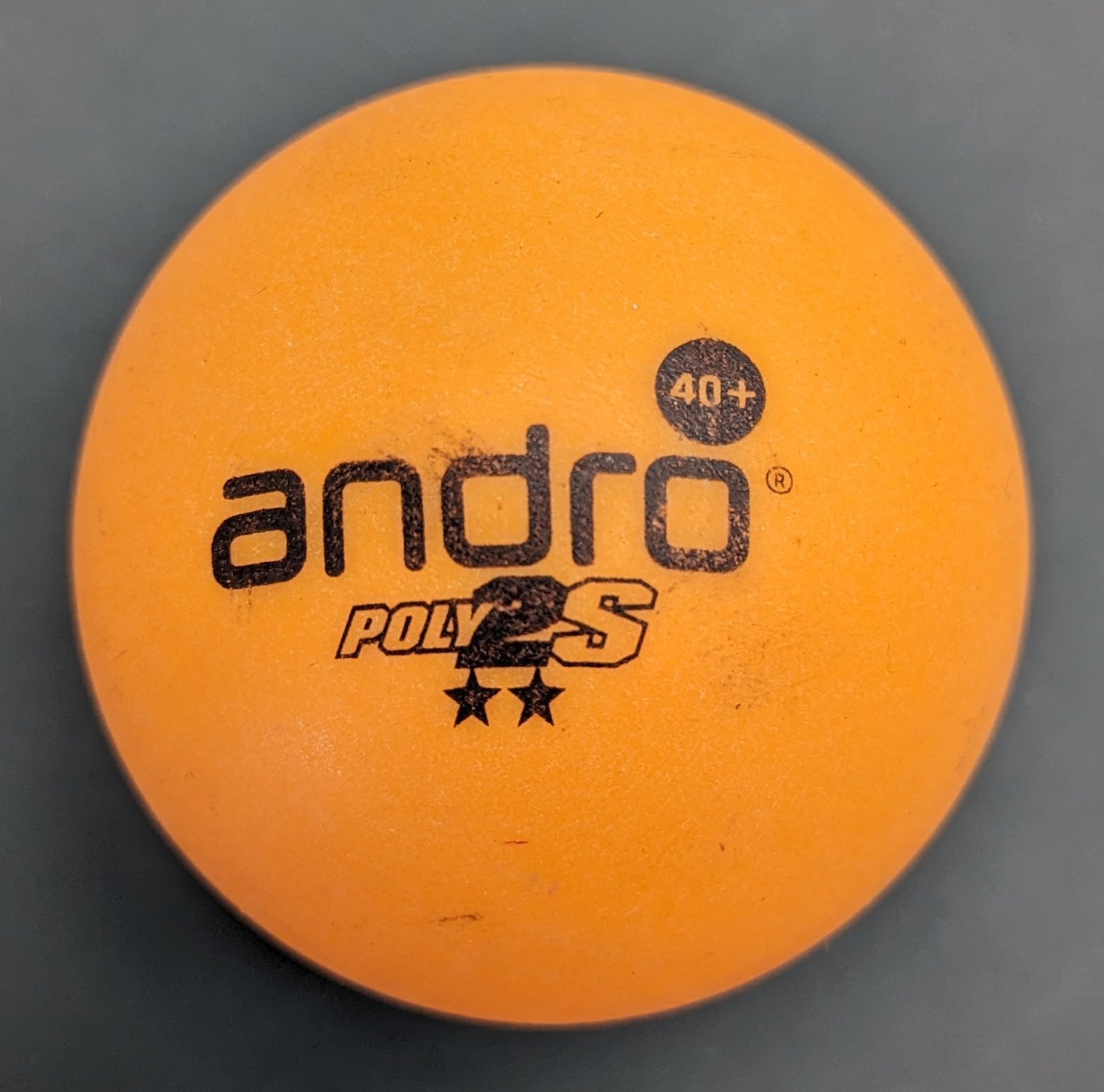}
    \caption{Logo 1}
   \label{fig:andro_logo}
  \end{subfigure}
  \hfill
  \begin{subfigure}{0.2\linewidth}
    \includegraphics[width=\linewidth]{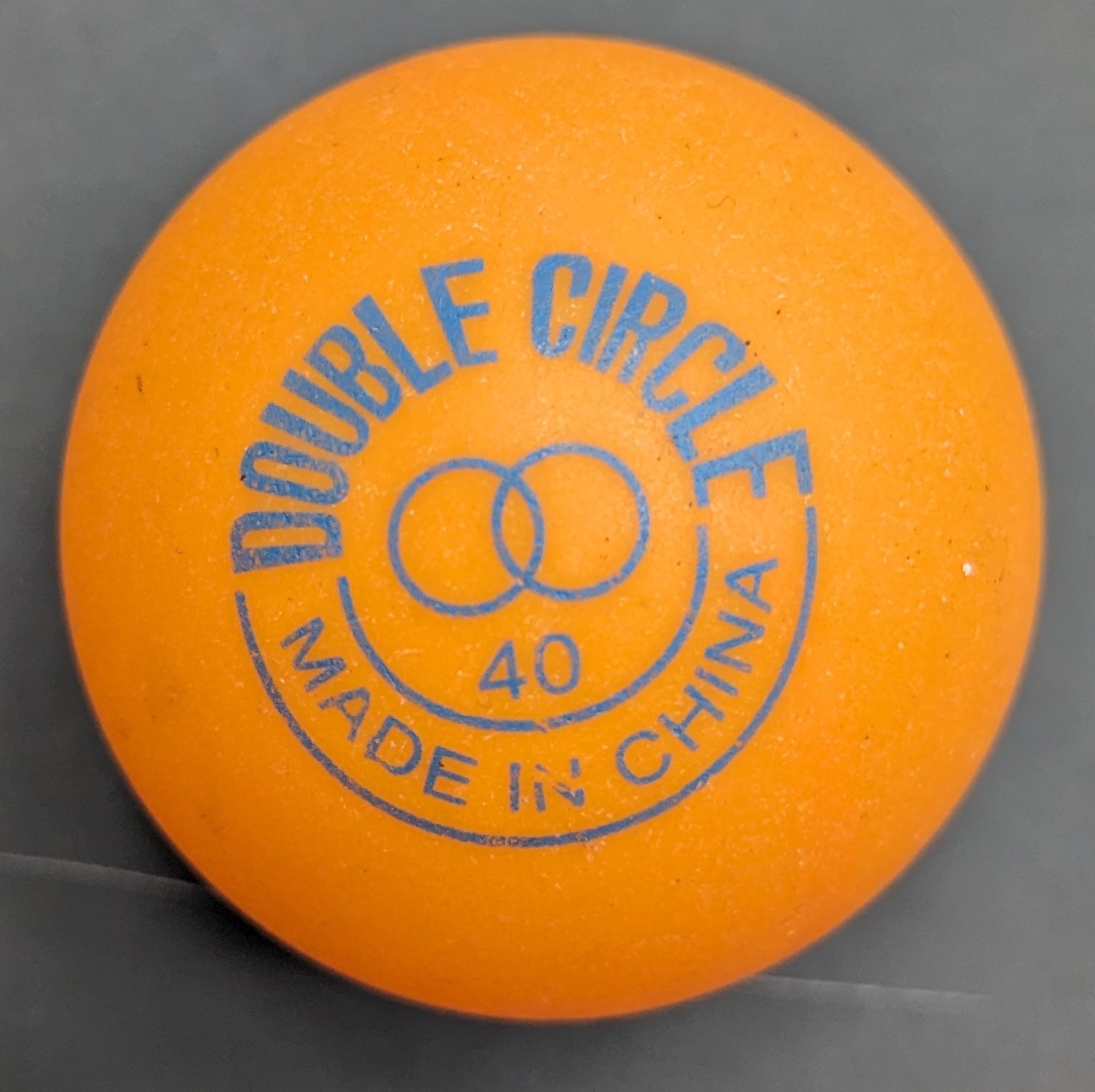}
    \caption{Logo 2}
   \label{fig:double_circle_logo}
  \end{subfigure}
  \hfill
  \begin{subfigure}{0.2\linewidth}
    \includegraphics[width=\linewidth]{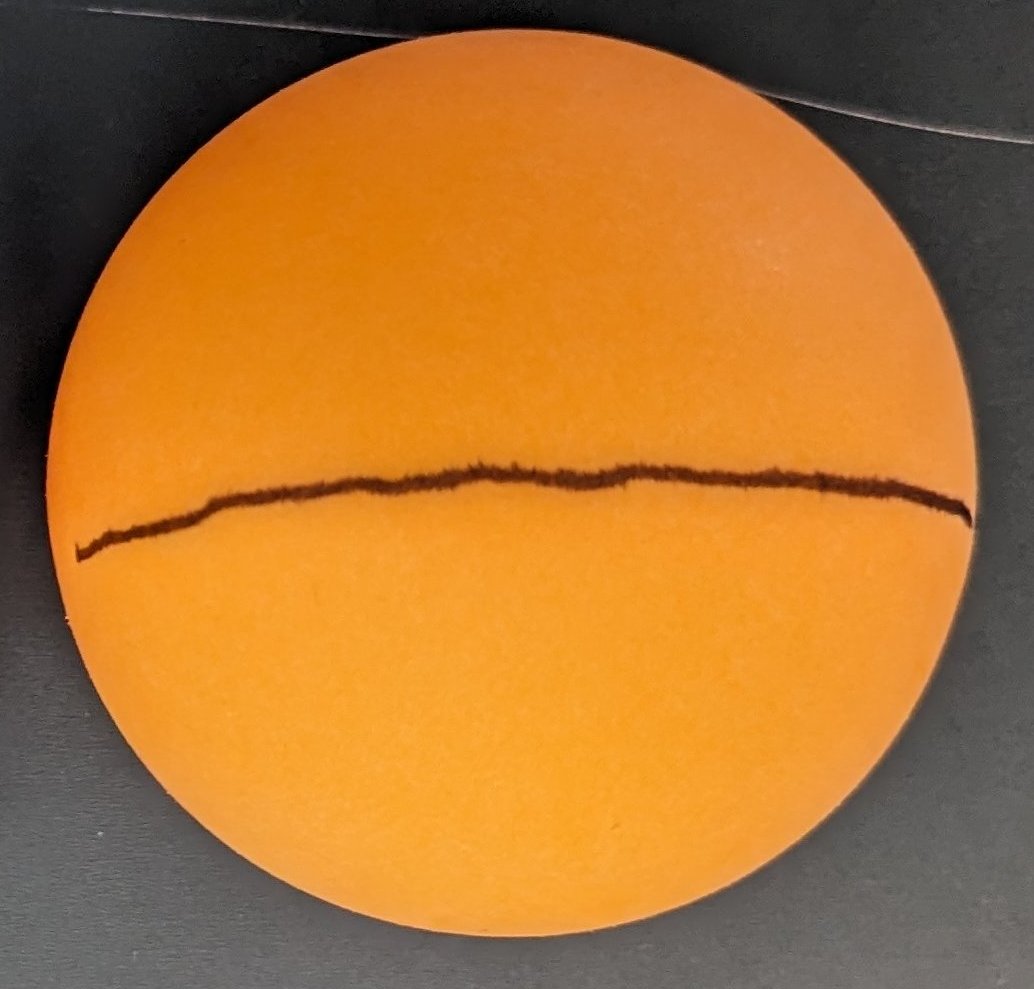}
    \caption{0.8mm line}
   \label{fig:line_ball_08}
  \end{subfigure}
  \hfill
    \begin{subfigure}{0.2\linewidth}
   \includegraphics[width=\linewidth]{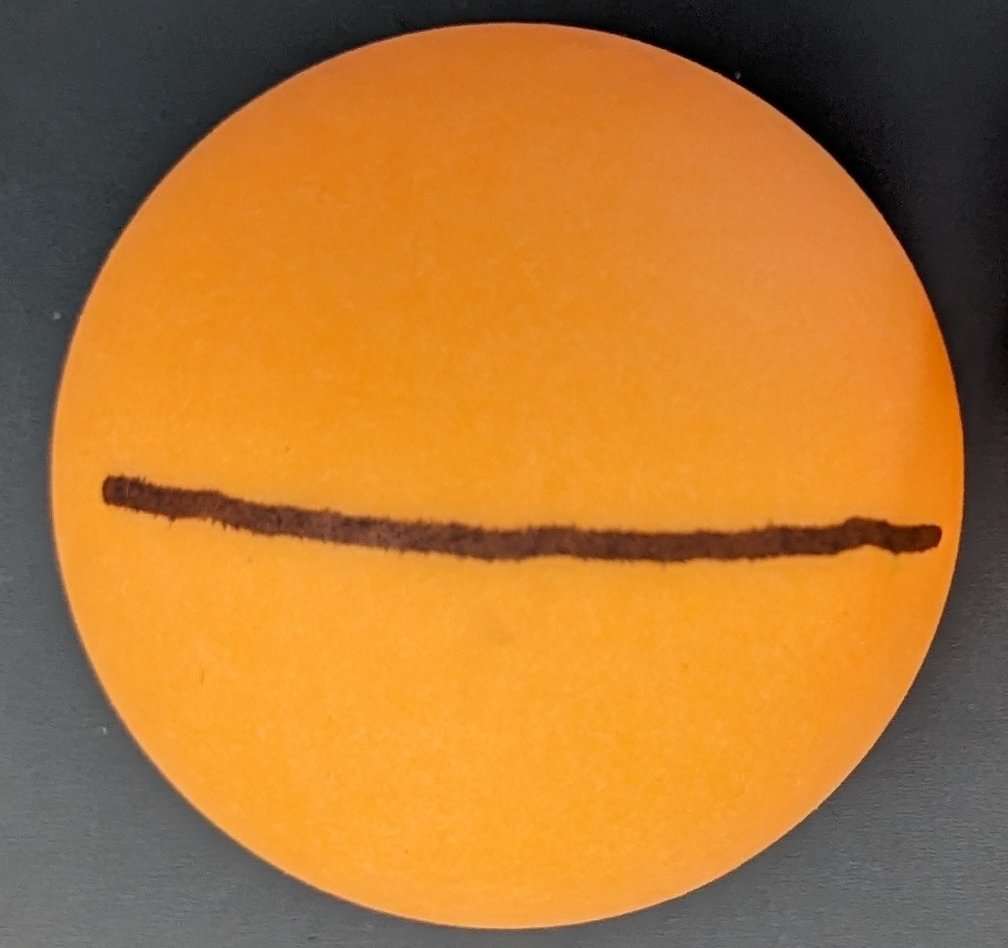}
   \caption{1.4mm line}
   \label{fig:line_ball_14}
  \end{subfigure}
  \caption{Different ball logos used to estimate the spin of the ball}
\end{figure}

While working on the spin estimation for the flying balls from a ball thrower, we noticed fewer events being generated compared to the ball spinner setup.
This led to an increase in spin estimation failures because there were too few events generated by the logo for the optical flow estimation.
Our interpretation of this phenomenon is that pixels respond slower to negative luminosity changes, such as the movement of a logo, after being triggered by a positive change, such as the appearance of the leading edge of a flying ball.
This is sometimes referred to as the event camera equivalent of motion blur.
Though our method works with the Andro logo displayed in~\cref{fig:andro_logo}, it does not generalize to all logos.
For example, our spin estimation algorithm does not work with the ball in~\cref{fig:double_circle_logo} since the logo does not generate enough events.
This is despite the fact that it works with the ball spinner.
We tested different custom logos to determine the causing factor.
The logo lines must be thicker than a certain threshold to be detected.
We drew lines of different line widths on logoless balls to show this.
The ball with the $0.8$mm line shown in~\cref{fig:line_ball_08} did not generate enough events, but the ball with $1.4$mm line shown in~\cref{fig:line_ball_14} worked well with our spin estimation method.
As such, only logos with lines of a certain width can be used for our spin estimation with an event camera.
Of course, there are different solutions to circumvent this.
%
%
We could use an event camera with a higher resolution or a lens with a longer focal length.
However, with the latter solution, the camera's field of view would no longer cover the whole width of table.

We also noticed that the spin estimation has a singularity.
Indeed, out of all the balls shown in~\cref{fig:balls}, there is one logo position, ball 5, for which the spin estimation failed completely, whatever the spin type.
Ball 5 does not work for backspin/topspin because the logo is barely visible on the side of the ball.
For sidespin, the estimation failed because the rotation axis and the logo coincided.
This could potentially be circumvented by using an optical flow algorithm better suited to our specific application.
\section{Conclusion}

In this work, we presented a spin estimation method for table tennis using optical flow based on the event stream of an event camera.
Though this method was implemented and tested for table tennis, it could be generalized to other sports.
We were able to compensate for the drawbacks of frame-based cameras with an event-based camera, which allows for more reliable spin estimation.

Although we have shown that our presented approach works for balls on a ball spinner and also for flying balls, depending on the setup, there are still some limitations, notably the slower reponse of OFF events of pixels after being triggered ON.
%

For future work, we plan to improve our setup and generate a dataset of event recordings of spinning balls to train a neural network specifically for event-based spin estimation.
This would allow us to bypass the optical flow estimation. 
With such a method, we hope to achieve faster and more accurate spin estimation.

{
    \small
    \bibliographystyle{ieeenat_fullname}
    \bibliography{main}
}

\clearpage
\setcounter{page}{1}
\maketitlesupplementary

\section{Camera settings}\label{sec:res_camera_settings}

The Prophesee EVK4 has five bias settings.
The name 'bias' comes from the electronics domain.
In electronics, 'biasing' usually refers to a fixed DC voltage or current applied to an electronic component, in order to establish proper operating conditions for the component.
However, despite the name, bias settings do not have to directly correlate with any voltage or current but is rather just a numerical value to adjust the property of the event camera.
The bias settings of the Prophesee EVK4 are listed in~\cref{tab:biases}.
On all the corresponding plots, the green line indicates our chosen setting.
\begin{table}[th]
    \centering
    \begin{tabularx}{\linewidth}{|c|X|}
        \hline
         $\textit{bias\_on} = 40$ & Contrast threshold for triggering on events \\
         \hline
         $\textit{bias\_off} = 40$ & Contrast threshold for triggering off events  \\
         \hline
         $\textit{bias\_fo} = 55$ & Cutoff frequency for the low-pass filter \\
         \hline
         $\textit{bias\_hpf} = 0$ & Cutoff frequency for the high-pass filter \\
         \hline
         $\textit{bias\_refr} = 80$ & Pixels' refractory period \\
         \hline
    \end{tabularx}
    \caption{
        List of the biases/settings of the event camera we used.
        Their tuned values (default values are $0$) and what they control (this is the naming convention used for Prophesee cameras)
    }
    \label{tab:biases}
\end{table}

To make the most out of event cameras, tuning the biases is essential to capture all the necessary events while maintaining a low level of noise.
This task is quite involved as the biases are interdependent.
%
%
Estimating the signal-to-noise ratio in a dynamic scene is also challenging since recreating the same scene for an objective comparison of the biases is not trivial.
Moreover, optimal biases are often task-specific and cannot be generalized. 
Automatic bias tuning was investigated in~\cite{Delbruck2021cvprw}. 
The authors used the event rate (ER) as the metric for finding the optimal biases, by tuning the biases independently.
With our setup, we have a static background and the only moving object in the scene is the table tennis ball.
This makes the ER a reliable metric for tuning the biases.
To recreate identical observations for different settings, we relied on a ball thrower to have the same ball trajectory every time.
However, the ball thrower can not guarantee the same logo position. 
For this reason, we used a robot arm to move the ball exactly the same way, with the required ball orientation.
This method, however, was only used for qualitative results (\cref{fig:logo_bias_on_off,,fig:logo_bias_refr}), as the moving robot arm influenced the event rate.

\subsubsection{\textit{bias\_on} and \textit{bias\_off}}\label{subsec:bias_on_and_bias_off}

Tuning the pixel sensitivity of the event camera with the biases \textit{bias\_on} and \textit{bias\_off} is a trade-off between having as little noise as possible while still capturing relevant events for our task. 
Increasing \textit{bias\_on} and \textit{bias\_off} will increase the contrast threshold and, therefore, decreases the sensor's sensitivity.

\begin{figure}[htb!]
  \centering
  \begin{subfigure}{\linewidth}
    \includegraphics[width=\linewidth]{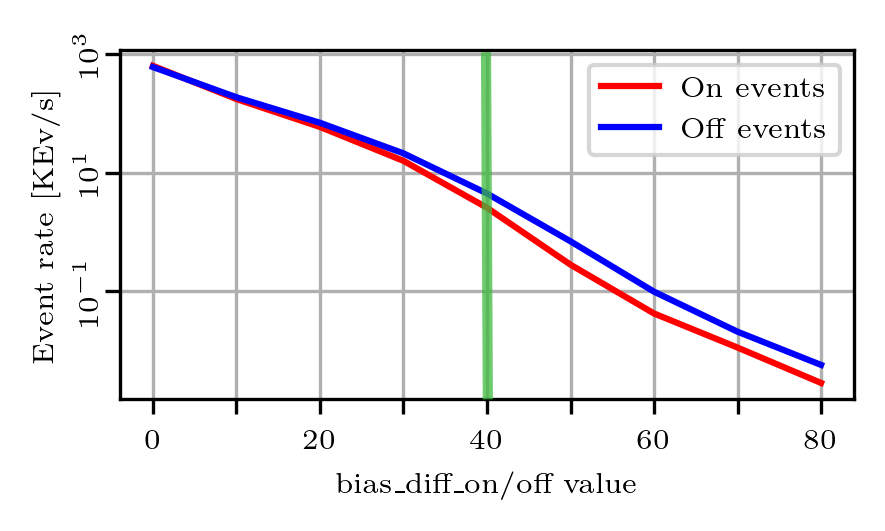}
    \caption{
            Event-rate for different \textit{bias\_on} and \textit{bias\_off} values while observing a static scene, where events can be considered to be noise.
            The y-axis is in log-scale.
    }
   \label{fig:er_noise_bias_on_off}
  \end{subfigure}
  \hfill
    \begin{subfigure}{\linewidth}
   \includegraphics[width=\linewidth]{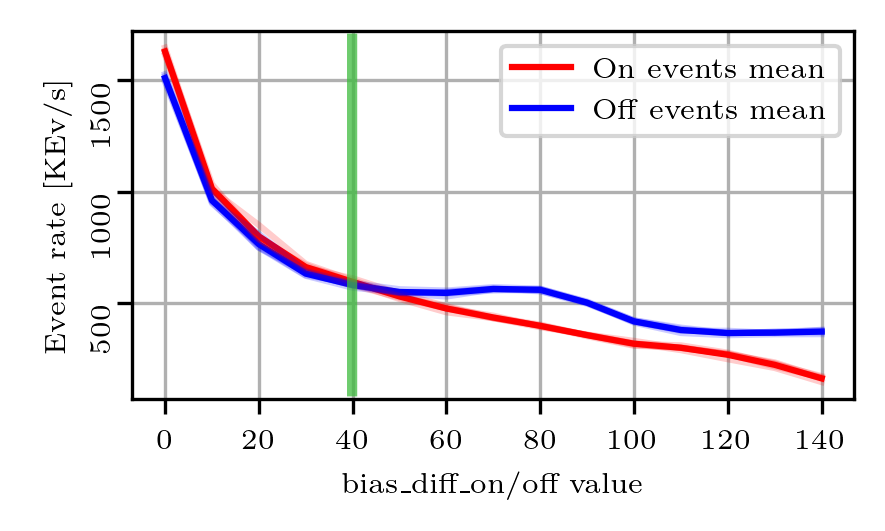}
   \caption{Event-rate for different \textit{bias\_on} and \textit{bias\_off} values while observing a flying ball (averaged over 5 samples)}
   \label{fig:off_on_ev_bias_off_on_ball}
  \end{subfigure}
    \caption{
            Event-rate for different pixel sensitivity (\textit{bias\_on} = \textit{bias\_off})
    }
\end{figure}

\Cref{fig:er_noise_bias_on_off} displays the event-rate for a static scene with regard to \textit{bias\_on} and \textit{bias\_off}.
As the scene is completely static, the generated events can be considered noise.
The noise is decreased to an acceptable level starting when \textit{bias\_on/off} goes over 40.

We also measured the event rate when a ball was flying in front of the camera.
We noticed that increasing \textit{bias\_on/off} further led to an unequal number of ON/OFF events when observing the ball, as shown in~\cref{fig:off_on_ev_bias_off_on_ball}.
This is due to the none symetrical behavior of the event pixels for OFF and ON events.
To avoid this behaviour, \textit{bias\_on/off} higher than 60 are to be avoided.
As such, we decided to set the \textit{bias\_on/off} to 40.

\begin{figure}[htb!]
  \centering
  \begin{subfigure}{0.3\linewidth}
    \includegraphics[width=\linewidth]{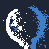}
    \caption{$\textit{bias\_on/off} = 40$}
  \end{subfigure}
  \hfill
  \begin{subfigure}{0.3\linewidth}
    \includegraphics[width=\linewidth]{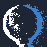}
    \caption{$\textit{bias\_on/off} = 90$}
  \end{subfigure}
  \hfill
\begin{subfigure}{0.3\linewidth}
    \includegraphics[width=\linewidth]{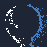}
    \caption{$\textit{bias\_on/off} = 140$}
  \end{subfigure}
  \caption{Accumulated event frames of the moving ball with the logo on the edge for different bias\_on/off values (with an accumulation time of $3\text{ms}$)}
  \label{fig:logo_bias_on_off}
\end{figure}

In~\cref{fig:logo_bias_on_off}, we show the effects of different \textit{bias\_on/off} on the accumulated event frame. 
We can clearly see a decrease in the number of generated events for increasing \textit{bias\_on/off}.


\subsubsection{\textit{bias\_fo}}\label{subsec:bias_fo}

\begin{figure}[htb!]
  \centering
  \begin{subfigure}{0.48\linewidth}
    \includegraphics[width=\linewidth]{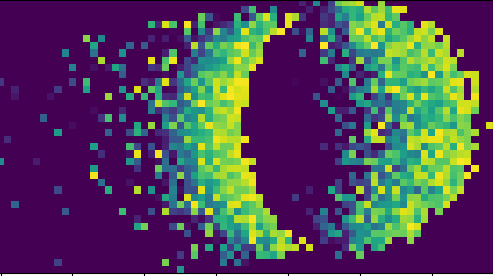}
    \caption{$\textit{bias\_fo} = -35$}
    \label{fig:ts_bias_fo_a}
  \end{subfigure}
  \hfill
  \begin{subfigure}{0.48\linewidth}
    \includegraphics[width=\linewidth]{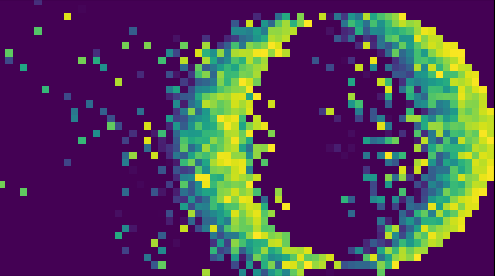}
    \caption{$\textit{bias\_fo} = 55$}
    \label{fig:ts_bias_fo_b}
  \end{subfigure}
  \caption{Linear time-surfaces ($\tau = 1ms$) of a flying ball with different \textit{bias\_fo}.}
  \label{fig:ts_bias_fo}
\end{figure}

\begin{figure}[htb!]
  \centering
   \includegraphics[width=\linewidth]{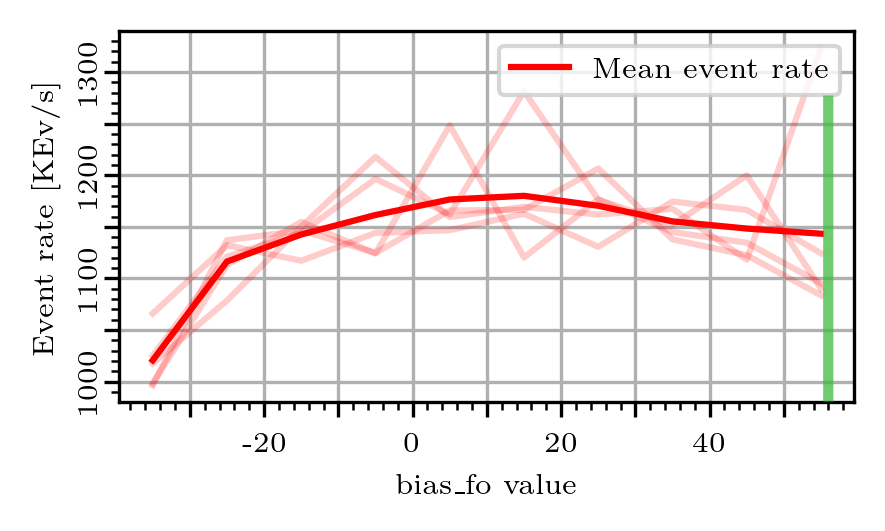}
   \caption{Event-rate for different \textit{bias\_fo} values while observing a flying ball (averaged over 5 samples)}
   \label{fig:er_bias_fo}
\end{figure}

The \textit{bias\_fo} controls the pixel's low-pass cut-off frequency.
It can filter out events generated by fast motions and flickering.
Increasing \textit{bias\_fo} will also increase the cut-off frequency.
There is a trade-off between noise and latency.
Indeed, a low-pass filter with a lower cut-off frequency will also introduce higher latency to the events as they will be triggered with some delay.
This can be observed in~\cref{fig:ts_bias_fo_a} where more new events are being triggered inside the ball.
These events are “late” events generated by the edge of the ball.
This can also be observed by the less distinct edge of the ball for $bias\_fo=-35$.
To avoid such “late” events being mixed up with events from the logo, the \textit{bias\_fo} was set to 55, which increased the cut-off frequency to the maximum.
However, from the event rate for different \textit{bias\_fo} values shown in~\cref{fig:er_bias_fo}, we see that this setting only affects the event rate for negative \textit{bias\_fo} values.
So any value between $0$ and $55$ would also work.

\subsubsection{\textit{bias\_hpf}}\label{subsec:bias_hpf}

\begin{figure}[htb!]
  \centering
   \includegraphics[width=\linewidth]{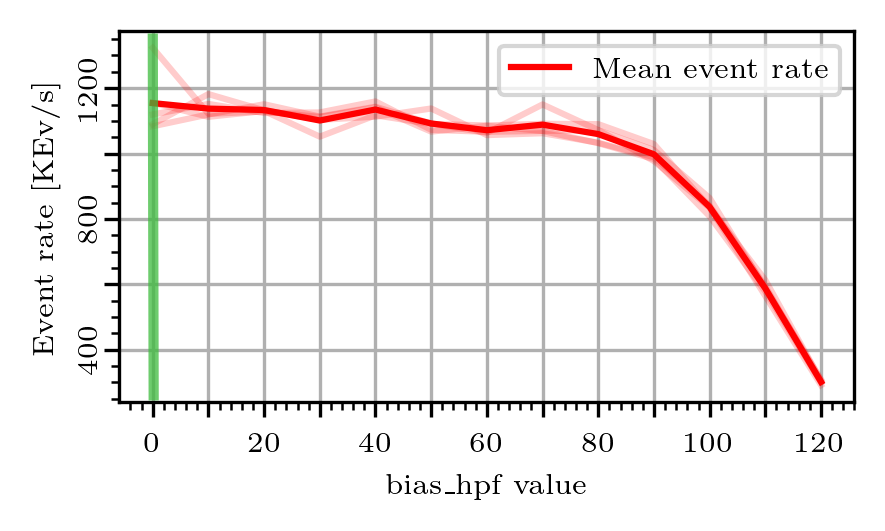}
   \caption{Event-rate for different \textit{bias\_hpf} values while observing a flying ball (averaged over 5 samples)}
   \label{fig:er_bias_hpf}
\end{figure}

The \textit{bias\_hpf} controls the pixel's high-pass cut-off frequency.
It allows filtering out low-frequency events such as noise and events generated from slow motions.
The event rate for different \textit{bias\_hpf} is shown in~\cref{fig:er_bias_hpf}.
We set \textit{bias\_hpf} to the minimum of $0$ for filtering out the least number of events.
All in all, the settings for \textit{bias\_fo} and \textit{bias\_hpf} were chosen to filter out the minimum amount of events.

\subsubsection{\textit{bias\_refr}}\label{subsec:bias_refr}

\begin{figure}[htb!]
    \centering
    \begin{subfigure}{\linewidth}
        \includegraphics[width=\linewidth]{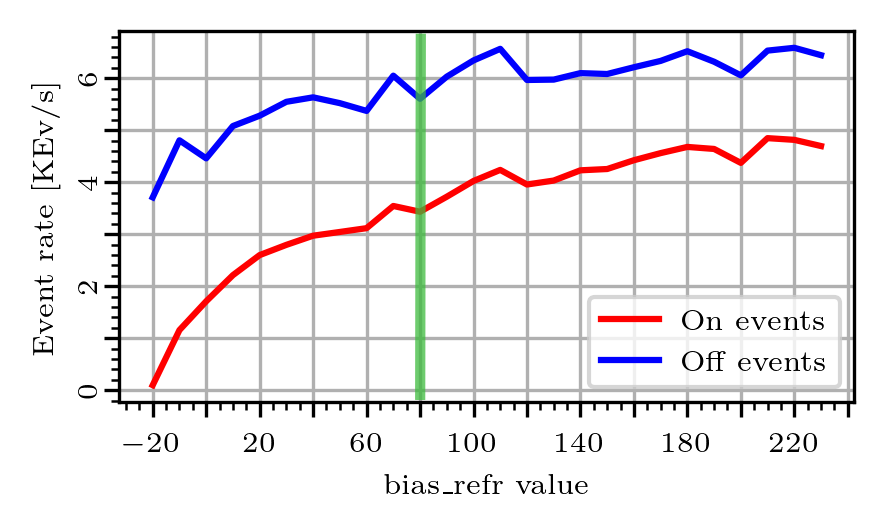}
        \caption{Event-rate for different \textit{bias\_refr} values while observing a static scene, where events can be considered to be noise}
        \label{fig:er_bias_refr_noise}
    \end{subfigure}
    \hfill
    \begin{subfigure}{\linewidth}
        \includegraphics[width=\linewidth]{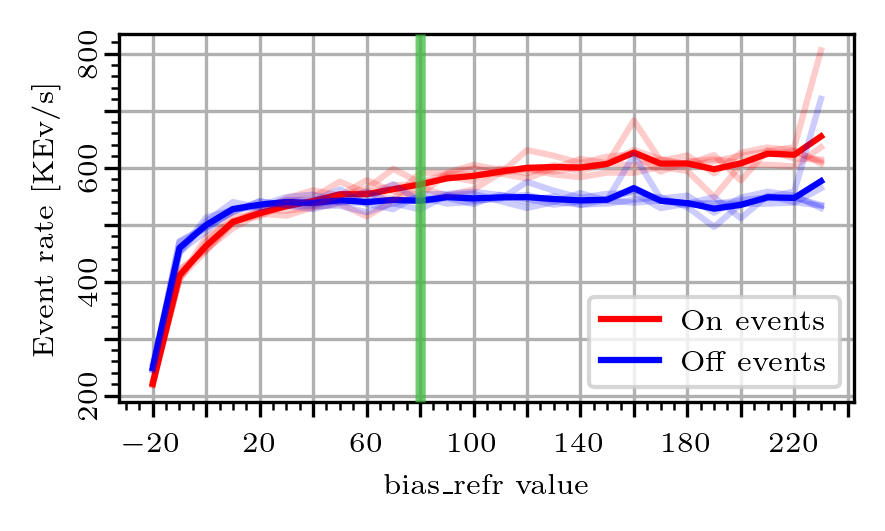}
        \caption{Event-rate for different \textit{bias\_refr} values while observing a flying ball (averaged over 5 samples)}
        \label{fig:er_bias_refr_ball}
    \end{subfigure}
    \caption{Event-rate for different \textit{bias\_refr} values}
    \label{fig:er_bias_refr}
\end{figure}

The \textit{bias\_refr} controls the pixel's refractory time, which is the time during which a pixel does not detect any change in illumination after it emitted an event.
Decreasing the pixel's refractory period will generate more events for a large illumination change.
A shorter refractory period will also lead to a higher event rate.
The event-rate for different \textit{bias\_refr} is shown in~\cref{fig:er_bias_refr}.
%

\begin{figure}[htb!]
  \centering
  \begin{subfigure}{0.3\linewidth}
    \includegraphics[width=\linewidth]{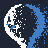}
    \caption{$\textit{bias\_refr} = 120$}
    \label{fig:logo_bias_short}
  \end{subfigure}
  \hfill
  \begin{subfigure}{0.3\linewidth}
    \includegraphics[width=\linewidth]{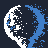}
    \caption{$\textit{bias\_refr} = 80$}
    \label{fig:logo_bias_refr_middle}
  \end{subfigure}
  \hfill
  \begin{subfigure}{0.3\linewidth}
    \includegraphics[width=\linewidth]{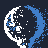}
    \caption{$\textit{bias\_refr} = -20$}
    \label{fig:logo_bias_refr_long}
  \end{subfigure}
  \caption{
  Accumulated event frames of the moving ball with the logo on the edge for different \textit{bias\_refr} values (with an accumulation time of $2\text{ms}$).
  A shorter refractory period of $120$ results in too many events/noise on the edge.
  On the other hand, a longer refractory period of $-20$ results in too few events for the spin estimation to work accurately.
  Therefore, we used a value of $80$, as a trade-off.
  }
  \label{fig:logo_bias_refr}
\end{figure}

We increased \textit{bias\_refr} to $80$ to not miss any events, which is also suggested by the camera's documentation.
We do not want the pixel to be “dead” at the ball's edge when the logo comes into view.
Thus, as soon as the edge moves to another pixel, that pixel should be able to trigger new events.
Since the nominal velocity of the flying ball is around $6000\; \text{pixels/s}$, the edge will move to another pixel in approximately every $0.1\text{ms}$.
Increasing \textit{bias\_refr} to values higher than 80, i.e., shortening the refractory period results in an increase in noise, as shown in~\cref{fig:er_bias_refr} where the ON event rate keeps increasing with \textit{bias\_refr}.
On the other hand, a higher refractory period results in too few events for the spin estimation to work accurately~\cref{fig:logo_bias_refr_long}.


\section{Filters}

In addition to the camera's bias, filters can be applied to the event stream to cancel redundant information or filter out noise.
The two options we considered are the Spatio-Temporal-Contrast (STC) filter and the TRAIL filter, as well as their combination.
The STC filter filters out isolated events that are not followed by other events of the same polarity. 
It does so by only retaining the second event from a burst of events.
The time window during which the second event is waited for can be tuned.

The TRAIL filter, as its name implies, gets rid of events that happen “behind” a moving edge with a certain time window, except if the event is of the opposite polarity.
These filters help clean the event stream by reducing noise and redundant information.
Both filters can be combined depending on the objective, only leaving one event out of a burst, giving much cleaner edges.
It should be noted that the filter's time window parameter should be tuned depending on the ball's velocity.

\begin{figure}[htb!]
    \centering
    \includegraphics[width=\linewidth]{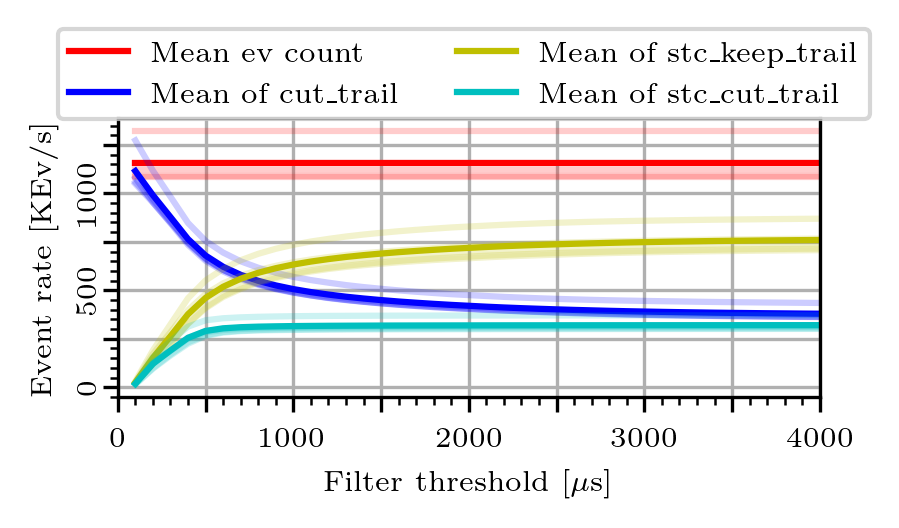}
    \caption{Effect of filter threshold on the event rate of a flying ball for different filter options}
    \label{fig:filter_rate}
\end{figure}

In~\cref{fig:filter_rate} we show how increasing the filter threshold affects the event rate of a flying ball.
For the STC filter, we keep very few of the observed events at low threshold values, as there are not enough events that are generated in quick enough succession to count as one burst. 
As the filter threshold increases, we still only keep a portion of the overall events, as for each burst (successive events at the same location), the filter discards the first one.
For the TRAIL filter, the opposite is true.
For each burst, only the first event is kept, and subsequent events within the threshold are discarded if they have the same polarity.
Thus, with a higher threshold, fewer events are kept.
Applying the STC and then the TRAIL filter (stc-cut-trail) keeps only one event of the same polarity for each burst.
Due to this, stc-cut-trail discards the most events but leads to the cleanest edges.

\begin{figure}[htb!]
    \centering
    \includegraphics[width=0.8\linewidth]{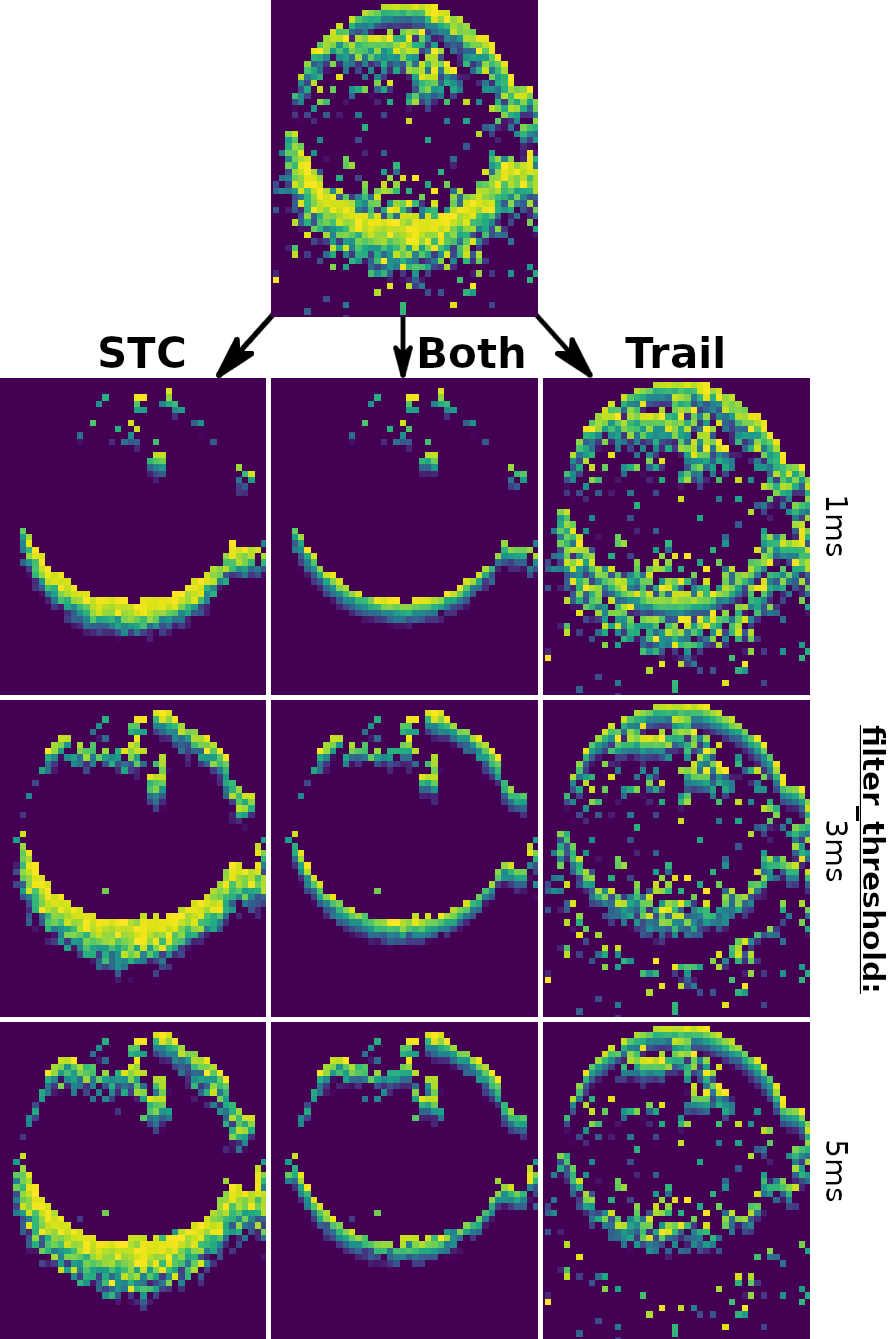}
    \caption{
            Time surfaces (positive and negative events, $5$ms accumulation time), showing the effect of the filter threshold on events kept by the different filter combinations.
            Top row: no filter.}
    \label{fig:filter_effects}
\end{figure}

\cref{fig:filter_effects} shows the events that the different filters keep depending on their threshold parameter, with the unfiltered version at the top.
For the STC filter, on low threshold values, very few events are being kept, as not enough events fall into the threshold time.
Especially for the front edge (top in image), we can see that it takes a higher threshold value before enough events are kept to make it out clearly.
It is worth noting that when the logo follows the front edge closely (as in the figure), the front edge will sometimes disappear. 
It is not entirely clear why this happens.
However, we believe it may be because not enough events from the front edge are generated before the logo's events are triggered.
Due to this, the STC filter removes the thin line of ON-events from the ball's edge, as it is treated as the start of the burst of the logo edge.
As clearly visible, the STC filter removes a lot of noise from the recording, as noise events are usually isolated and not part of a burst of events generated by a moving edge.
However, behind the initial edge, there are still a lot of trailing events.

The TRAIL filter's effects are very clearly visible as the threshold value increases, especially at the back (bottom) of the ball.
After an event from the edge is recorded, subsequent events of the same polarity that fall into the threshold time, are removed.
This means that many trailing events generated by the edge are cut out.
We can clearly see the gap behind the back edge increase as the filter threshold increases.
Note, however, that the logo is not being removed behind the front edge, as its polarity change resets the filter window.

Combining the STC and trail filter leaves us with a very clear edge and little remaining noise.
Increasing the threshold over 5000$\mu$s changed very little about the number and quality of events kept, so we use this as our $filter\_threshold$ when applying STC-cut-trail filter.

\section{Optical flow estimation}
\begin{figure}[htb!]
    \centering
    \includegraphics[width=0.6\linewidth]{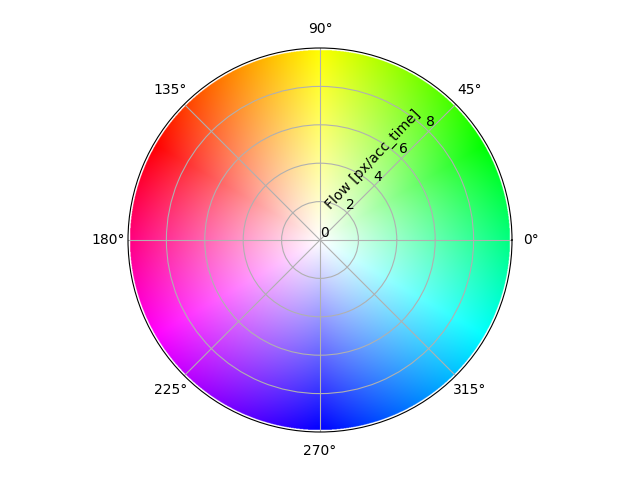}
    \caption{HSV color wheel used to represent the optical flow}
    \label{fig:hsv_wheel}
\end{figure}

We represent the flow using the HSV color scheme for all the optical flow estimations in~\cref{fig:hsv_wheel}, where high saturation indicates high flow, and the hue indicates the flow direction.

\begin{figure}[htb!]
    \centering
    \includegraphics[width=\linewidth]{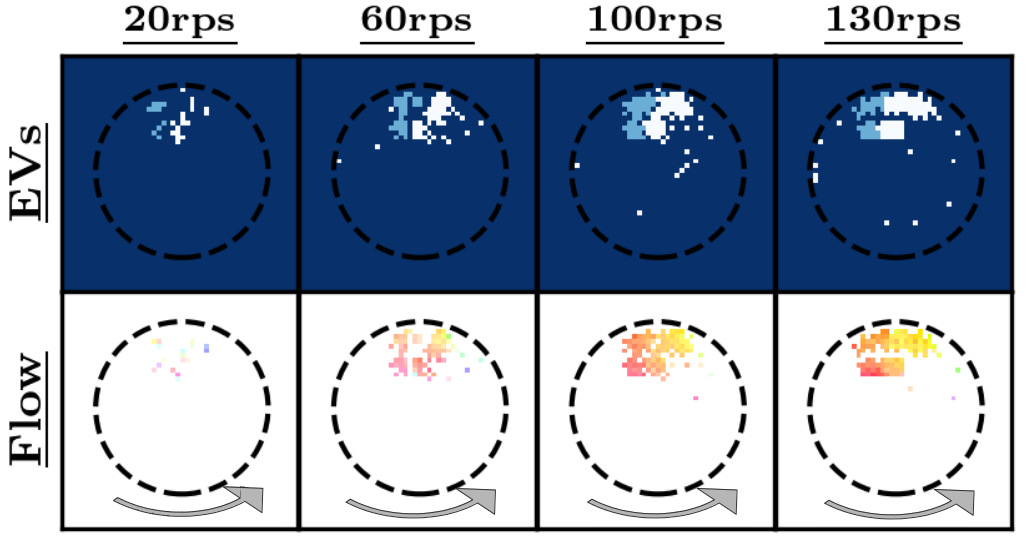}
    \caption{
            Optical flow for sidespin with $t_{acc}= 0.715 ms$.
            The arrows indicate the direction of the spin.
    }
    \label{fig:res_optical_flow_sidespin}
\end{figure}

In~\cref{fig:res_optical_flow_sidespin}, we show examples of accumulated event frames and flow estimates of sidespin for different spin values.
As clearly visible, higher spin values lead to more events, allowing more accurate flow estimation.

\begin{figure}[htb!]
    \centering
    \includegraphics[width=\linewidth]{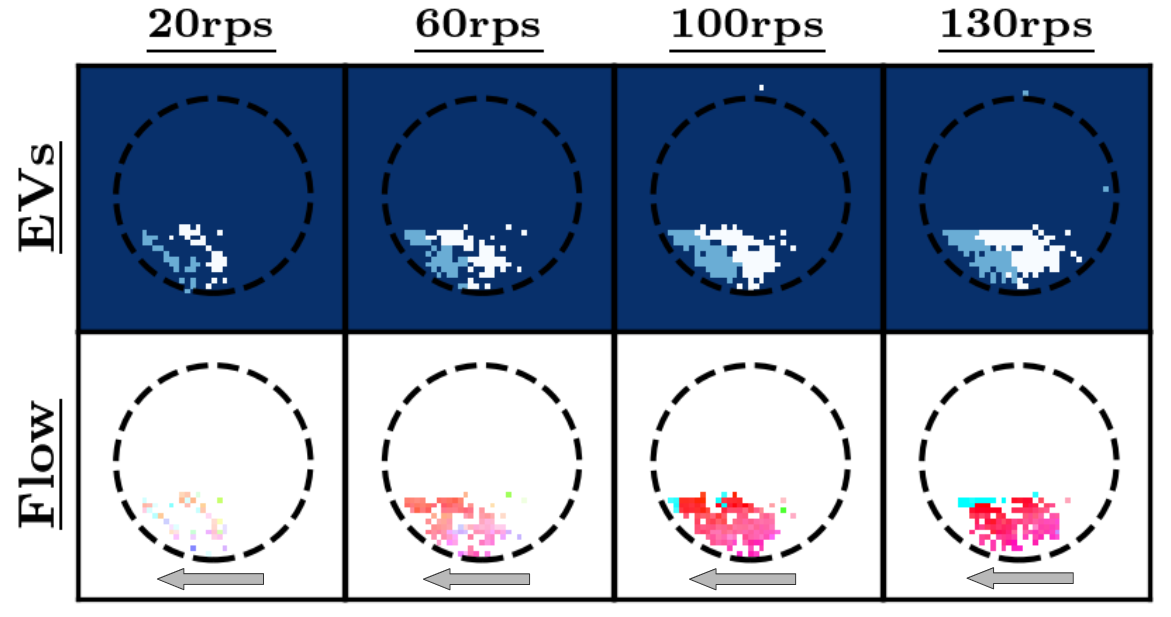}
    \caption{
            Optical flow for backspin/topspin with $t_{acc}= 0.715 ms$.
            The arrows indicate the direction of the spin.
    }
    \label{fig:res_optical_flow_v_spin}
\end{figure}

Examples of accumulated event frames and corresponding flow estimates of backspin/topspin for different spin values are visualized in~\cref{fig:res_optical_flow_v_spin}.
The same as for the sidespin, higher spin values result in more events and, thus, more accurate optical flow.

\section{Cleaning the EROS time surface}\label{sec:supp_cleaning_eros}

\begin{figure}[htb!]
    \centering
    \begin{subfigure}{\linewidth}
        \includegraphics[width=\linewidth]{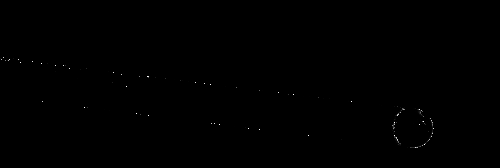}
        \caption{No cleaning}
        \label{fig:no_clean}
    \end{subfigure}
    \begin{subfigure}{\linewidth}
        \includegraphics[width=\linewidth]{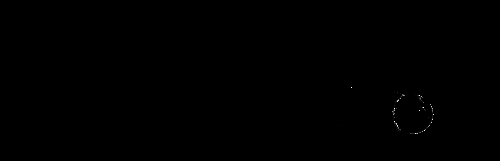}
        \caption{With cleaning}
        \label{fig:hit_or_miss}
    \end{subfigure}
    \caption{The EROS time surface (a) without cleaning and (b) with cleaning}
\end{figure}

Even with the STC filter, noise makes it through to the EROS time surface.
Not only noise can pollute the time surface, but also trail events, as seen in~\cref{fig:no_clean}.
To avoid misdetecting circles with the Hough Transform, we eliminate these isolated events with a hit-or-miss detection.
The EROS time surface without cleaning is shown in~\cref{fig:no_clean} and with cleaning in~\cref{fig:hit_or_miss}.

The kernel we run for the hit-and-miss is defined as
\begin{equation}
    k = \begin{bmatrix}
    -1 & -1 & -1 & -1 \\
    -1 & 1 & 1 & -1 \\
    -1 & 1 & 1 & -1 \\
    -1 & -1 & -1 & -1 
    \end{bmatrix}.
\end{equation}

\section{Generating ground truth for the ball detector benchmark}

\begin{figure}[htb!]
    \centering
    \includegraphics[width=0.6\linewidth]{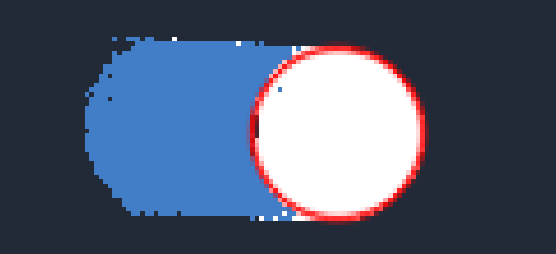}
    \caption{
            Example of ball position labeling with blob detection.
            The red circle is the ball detected by the blob detection.
    }
    \label{fig:labeling_acc_frame}
\end{figure}

We rely on event accumulation frames and blob detection to automatically generate ground truth for the ball detection.
To do so, we use a large accumulation time of $t_{acc} = 10\text{ms}$ to generate accumulated event frames.
This makes the ball clearly visible for blob detection, as shown in~\cref{fig:labeling_acc_frame}.

\section{Ball thrower benchmark}
To capture the same ball trajectories, the event and frame cameras were installed next to each other.
Despite trying to align them as best as possible, it is not possible for them to exactly share the same field of view. 
We thus needed to calculate the transformation from the event camera to the frame camera for the benchmark.
The transformation between both cameras was calculated in the same fashion as for a stereo camera setup.
A chessboard pattern displayed on a LCD screen was used for that purpose.
The screen was set to blink for the event camera and to static display for the frame camera.
Thanks to this, the spin calculated with the event camera could be transformed to match the spin calculated with the frame camera.

Regarding the generation of the ground truth, SpinDOE~\cite{Gossard2023iros} was used.
It can indeed estimate the table tennis ball spin with high accuracy.
It has a relative error of $1\%$ for the spin magnitude and a spin axis error of $2.4\degree$.
However, the ball thrower used has some stochasticity to its behaviour: even though the settings are the same, the ball will not have the same trajectory.
To compensate for this, dot-patterned balls were shot 5 times with the same settings for the ball thrower and the ground truth is assumed to be the mean spin vector. 

\begin{table*}[!ht]
    \begin{subtable}{1\textwidth}
    \centering
    \begin{tabular}{c|ccccccccccccc}
         & -5 & -4 & -3 & -2 & -1 & 0 & 1 & 2 & 3 & 4 & 5 & 6 & 7 \\
        \hline
        25 & 137.81 & 124.60 & 107.42 & 103.23 & 98.22 & 81.83 & 63.69 & 38.87 & 35.72 & 41.38 & 62.41 & - & -\\
        20 & 127.10 & 110.70 & 92.45 & 93.09 & 88.47 & 72.36 & 60.61 & 45.53 & 45.40 & 53.64 & 64.80 & - & -\\
        15 & 119.51 & 98.76 & 81.94 & 81.89 & 72.69 & 66.75 & 63.80 & 57.20 & 57.48 & 63.80 & 68.69 & - & -\\
        10 & 92.31 & 74.41 & 60.74 & 61.23 & 58.74 & 57.85 & 61.46 & 61.80 & 62.27 & 65.54 & 66.11 & - & -\\
    \end{tabular}
    \caption{Sidespin}
    \end{subtable}
    \bigskip
    \begin{subtable}{1\textwidth}
    \begin{tabular}{c|ccccccccccccc}
         & -5 & -4 & -3 & -2 & -1 & 0 & 1 & 2 & 3 & 4 & 5 & 6 & 7 \\
        \hline
        25 & 111.81 & 90.41 & 64.34 & 65.46 & 53.42 & 34.13 & 17.01 & 16.61 & 35.10 & 54.20 & 74.14 & - \\
        20 & 99.99 & 78.42 & 55.22 & 53.91 & 39.42 & 13.30 & 2.05 & 22.39 & 39.94 & 54.05 & 68.90 & - & - \\
        15 & 93.11 & 71.01 & 47.27 & 39.65 & 22.80 & 5.65 & 4.77 & 25.30 & 40.17 & 54.37 & 62.67 & 95.77 & - \\
        10 & 72.33 & 54.58 & 36.46 & 23.41 & 9.21 & 0.91 & 7.41 & 17.10 & 25.04 & 34.94 & 38.67 & 59.75 & 88.76 \\
    \end{tabular}
    \caption{Back/top-spin}
    \label{tab:gt_ball_thrower_back_top}
    \end{subtable}
    \caption{Ground truth values of the spin magnitude [rps] calculated with SpinDOE for the ball thrower. The rows represent the different velocity settings and the columns the spin settings.}
    \label{tab:gt_spins}
\end{table*}

In \cref{tab:gt_spins}, we list all the ground truth values for the different velocity and spin settings.
The ground thruth values for the higher spins could not always be calculated because of the motion blur due to high spins.
It should be noted that with the ball thrower used, the ball spin magnitude also increased with higher ball velocity settings.

\end{document}